%% file: main.tex
\documentclass[acmlarge,12pt]{acmart}
\usepackage{bm,amsmath,amsfonts}
\usepackage{caption}
\usepackage{subcaption}

\AtBeginDocument{%
  \providecommand\BibTeX{{%
    \normalfont B\kern-0.5em{\scshape i\kern-0.25em b}\kern-0.8em\TeX}}}

\usepackage{multirow,array}
\usepackage{float}
\usepackage{tikz}
\def\checkmark{\tikz\fill[scale=0.4](0,.35) -- (.25,0) -- (1,.7) -- (.25,.15) -- cycle;} 
\begin{document}

\title{On the Explainability of Natural Language Processing Deep Models}
\author{Julia El Zini}
\email{jwe04@aub.edu.lb}
\orcid{0000-0001-7499-8668}
\author{Mariette Awad}
\email{mariette.awad@aub.edu.lb}
\orcid{0000-0002-4815-6894}
\affiliation{%
  \institution{Department of Electrical and Computer Engineering, American University of Beirut.}
  \streetaddress{P.O. Box 11-0236, Riad El Solh}
  \city{Beirut}
  \state{Lebanon}
  \postcode{1107-2020}
}

\begin{abstract}
\input{abstract}
\end{abstract}



\maketitle

\section{Introduction}\label{sec:intro}
\input{intro}

\section{Challenges in ExAI on NLP Models}\label{sec:challenges}
\input{challenges}

\section{Related Surveys}\label{sec:lit}
\input{surveys}

\section{Terminology}\label{sec:terminology}
\input{terminology}



\section{Interpreting Word Embeddings}\label{sec:NLPembeddings}
\input{nlp_wordembeddings}

\section{What do RNNs Learn?}\label{sec:RNNs}
\input{rnn}

\section{What do Transformers Learn?}\label{sec:transformers}
\input{transformer}

\section{Explaining  Model's Decisions}\label{sec:NLPdecision}
\input{exp_decisions}



\section{Discussion and Future Directions}\label{sec:future}
\input{future}

\section{Conclusion}\label{sec:conc}
\input{conc}

\bibliographystyle{ACM-Reference-Format}
\bibliography{acmart}

\end{document}


\appendix
\section{Case-study on Neural Machine Translation (NMT)}
\input{nmt}

\section{Visualization Tools}
\input{vis}
\clearpage
\bibliographystyle{ACM-Reference-Format}
\bibliography{acmart}

%% file: abstract.tex
Despite their success, deep networks are used as black-box models with outputs that are not easily explainable during the learning and the prediction phases. This lack of interpretability is significantly limiting the adoption of such models in domains where decisions are critical such as the medical and legal fields. Recently, researchers have been interested in developing methods that help explain individual decisions and decipher the hidden representations of machine learning models in general and deep networks specifically. While there has been a recent explosion of work on \textbf{Ex}plainable \textbf{A}rtificial \textbf{I}ntelligence (\textbf{ExAI}) on deep models that operate on imagery and tabular data, textual datasets present new challenges to the ExAI community. Such challenges can be attributed to the lack of input structure in textual data, the use of word embeddings that add to the opacity of the models and the difficulty of the visualization of the inner workings of deep models when they are trained on textual data.

Lately, methods have been developed to address the aforementioned challenges and present satisfactory explanations on Natural Language Processing (NLP) models. However, such methods are yet to be studied in a comprehensive framework where common challenges are properly stated and rigorous evaluation practices and metrics are proposed. 

Motivated to democratize ExAI methods in the NLP field, we present in this work a survey that studies \textit{model-agnostic} as well as \textit{model-specific} explainability methods on NLP models. Such methods can either develop \textit{inherently} interpretable NLP models or operate on pre-trained models in a \textit{post-hoc} manner. We make this distinction and we further decompose the methods into three categories according to what they explain: (1) word embeddings (input-level), (2) inner workings of NLP models (processing-level) and (3) models' decisions (output-level). We also detail the different evaluation approaches interpretability methods in the NLP field. Finally, we present a case-study on the well-known neural machine translation in an appendix and we propose promising future research directions for ExAI in the NLP field.

%% file: intro.tex
Ever since their introduction, deep learning (DL) models are revolutionizing several NLP applications ranging from machine translation \cite{wu2016google,vaswani2018tensor2tensor,bahdanau2014neural} to text summarization \cite{tas2007survey} and question answering \cite{yang2016stacked,devlin2018bert,xu2016ask}. AI-powered systems that mainly use DL models are reaching highly accurate predictions that surpass human performances in some cases \cite{silver2016mastering}. 
However, due to their non-linear multilayered structure, deep networks are often seen as black-box models that achieve high performances; but in an opaque manner.  Researchers and practitioners often question how credible these predictions are if the reasoning behind them is a highly non-linear enigma that cannot be easily deciphered. This black-box nature of DL models gives rise to several criticisms of their non-transparent predictions. Transparency and interpretability are thus needed to establish user trust when black-box DL achieves a performance comparable to that of the human. Besides, weak DL models entail interpretability to investigate failure cases and direct the researchers in the proper paths \cite{ding2017visualizing}. The need for interpretability is even more pronounced when DL models beat human performance where explanations can serve as a machine teaching framework for the human to make better decisions. For instance, if the decision-making system of Alpha Go \cite{silver2016mastering} that beat the world champion in the Go game was transparent, some creative moves can be taught to humans to help them learn the game or even extend their mental capabilities.

Transparency is not only needed on the prediction level; some situations require DL models that achieve explainability during the learning phases. For instance, DL models which learn from curated datasets might engender bias that is not easy to detect requiring a higher explainability level \cite{mooney1996comparative}. Interpretability of the hidden representations of these networks and the understanding of the predictions of the whole category would suggest whether some protected attributes are affecting the predictions in a biased manner. 

Recently, the ExAI field has attracted researchers to develop explainability methods for black-box deep networks \cite{lundberg2017unified,ribeiro2016should,selvaraju2017grad} in general. While some of these methods can be directly applied to NLP models \cite{bach2015pixel,ribeiro2016should,lundberg2017unified}, others are specific to imagery datasets or Convolutional Neural Networks (CNNs) that are not very suitable for neural machine understanding tasks \cite{mahendran2015understanding,selvaraju2017grad,dosovitskiy2015inverting,dabkowski2017real}. It is worth mentioning that explainability and interpretability are used interchangeably in this work. For definitions and terminologies, readers are referred to \cite{miller2019explanation}.

datasets or Convolutional Neural Networks (CNNs) that are not very suitable for neural machine understanding tasks \cite{mahendran2015understanding,selvaraju2017grad,dosovitskiy2015inverting,dabkowski2017real}. We focus our survey on the interpretability of language models or deep networks that operate on textual data. We make the distinction between the methods that develop \textit{inherently} interpretable models and those that operate in a \textit{post-hoc} manner. We also make the distinction between \textit{model-specific} and \textit{model-agnostic} methods and the level at which each method operates: input- (or embedding), processing- and prediction-level.  Throughout this work, we use the terms explainability and interpretability interchangeably. For more details on definitions and terminologies, we refer readers to \cite{miller2019explanation}. We make the distinction between the methods that develop \textit{inherently} interpretable models and those that operate in a \textit{post-hoc} manner. We also make the distinction between \textit{model-specific} and \textit{model-agnostic} methods and the level at which each method operates: input- (or embedding), processing- and prediction-level. 

Although interpretability is a relatively new research track in AI, different surveys have been compiled to highlight specific interpretability aspects \cite{doshi2017towards,montavon2018methods,qin2018convolutional,gilpin2018explaining,nguyen2019understanding}. Recently, \cite{danilevsky2020survey} addressed interpretability of NLP models in their survey while focusing on explaining model's decision. Unlike \cite{danilevsky2020survey}, this work is not only limited to ExAI methods on the decision level. Our work here contributes to the community the first survey on the (1) interpretability of word embeddings models which constitute a crucial part of NLP, (2) the inner representations of NLP networks, and (3) the transformer models which presented a great debate on the interpretability of their attention mechanisms. In addition to that, we survey existing work on the explainability of individual model decisions with insights about research challenges and opportunities in that field. We also highlight different empirical setups, metrics, and datasets that NLP researchers rely on to evaluate their ExAI methods.

The rest of this paper is organized as follows: Section~\ref{sec:challenges} presents the challenges that general ExAI methods face with textual datasets or NLP models and Section~\ref{sec:lit} presents the related surveys to ExAI. Then, Section~\ref{sec:terminology} presents the terminology used in this paper with respect to the three proposed dimensions.  The rest of the sections are focused on the interpretability of NLP models on three levels: the input, the processing, and the output. Section~\ref{sec:NLPembeddings} surveys existing work on the interpretability of word embeddings serving as inputs to NLP models, Section~\ref{sec:RNNs}~and~\ref{sec:transformers} present the interpretability methods applied to the inner representations of Recurrent Neural Networks (RNNs) and transformers, with their debatable attention mechanisms, respectively during the processing phase and Section~\ref{sec:NLPdecision} focuses on the explanations of individual predictions or outputs. 
Finally, we end with some concluding remarks and future directions in Sections~\ref{sec:conc} and \ref{sec:future} while presenting a case study on neural machine translation in an appendix.

%% file: challenges.tex
In the traditional learning framework, textual datasets present many challenges such as polysemy, sarcasm, slang, cultural effect and ambiguity. Explainability methods proliferate these challenges. In what follows, we identify some text-specific challenges that hamper the application of general ExAI methods into NLP models. Those challenges motivate the necessity of addressing ExAI in NLP independently from other ExAI techniques.

First, some of the explainability methods provide their explanations in terms of specific input features, such as pixels. These features are not as straightforward in textual datasets. The majority of NLP models operate on embeddings that are opaque representations, as opposed to pixels or numerical values. Providing explanations in terms of specific embedding dimensions will not have a practical implication for the explanations. Additionally, the inner workings of deep networks trained on text cannot be easily visualized as opposed to visual networks. Further processing on the inner encodings is required to dissect the learned knowledge.

When a model's decision is explained by the words that contributed to the prediction, further refinement is needed. For instance, explaining decisions in terms of the input features can be easily formulated when the input is numerical or imagery where decisions can be reflected by clear features or pixels. This task becomes challenging with text where the syntactic and semantic features cannot be easily dissected in the input to interdependently serve as explanations. After all, a word is a fusion of syntax, semantics, and previous context. When providing the explanation as a set of words, one can thus inquire if the explanation model is attending to the part-of-speech tag, the entity, the meaning, or the accumulated context. 

Long-term dependencies, on the other hand, add to the challenges of ExAI methods for NLP models. Namely, textual explanations might not always be a set of consecutive words but words with longer dependencies as opposed to neighboring pixels in images. Finally, multi-lingual support for deep models language models specifically introduces new challenges to explainability where the encoded language semantics and context need further deciphering.


%% file: surveys.tex
Given the infancy of the ExAI field, especially in the context of NLP models, there is a handful of surveys that describe its terminology, taxonomy, different methods, and evaluation frameworks. Doshi et al. introduce the taxonomy of ExAI in \cite{doshi2017towards} while setting the common ground for rigorous evaluation of interpretability of machine learning models through application-grounded, human-grounded, and functionally-grounded settings. Montavon et al. \cite{montavon2018methods} focus on activation maximization techniques, sensitivity analysis, Taylor decomposition, and relevance propagation. 
Their work is specific to \textit{post-hoc} interpretability methods and does not discuss \textit{inherently} transparent models.  A subset of the interpretability methods is also surveyed in \cite{qin2018convolutional}, where a comprehensive study is presented covering activation maximization, network inversion, deconvolutional neural networks, and network dissection based visualization applied on imagery datasets. Similarly, Guidotti et al. \cite{guidotti2018survey} focus on decision rules, features importance, saliency masks, sensitivity analysis, partial dependence plot, prototype selection, and neurons activation methods while mainly studying image and tabular datasets. 

In \cite{gilpin2018explaining}, a brief survey that discusses linear proxy models, decision trees, automatic rules, and saliency maps, is presented. This survey studies a few approaches that are \textit{inherently} explainable such as attention networks, and disentangled representations that are designed to generate explanations. Later, Nguyen et al. \cite{nguyen2019understanding} discuss how the activation maximization approaches evolved. Their survey is limited to imagery datasets and is very specific to optimization techniques in activation maximization. The survey of \cite{tjoa2019survey} is also specific to explainability methods in the medical field, which implies signals and imagery datasets. Very recently, \cite{das2020opportunities} introduce the opportunities and challenges of ExAI while discussing the limitations of certain methods and \cite{puiutta2020explainable} present ExAI methods applied to reinforcement learning settings.

The majority of the surveys in the literature are either brief \cite{dovsilovic2018explainable,gilpin2018explaining} or are focused on imagery datasets \cite{qin2018convolutional,das2020opportunities,nguyen2019understanding}. It is not until late 2020 that a survey on ExAI integrated with NLP models has been introduced in \cite{danilevsky2020survey}. In this survey, Danilevsky et al. categorized the explanation types and methods while focusing on visualizing techniques and presenting some gaps and research direction in the ExAI-NLP area. 
A few months later, a survey on explanation-based human debugging of NLP models\cite{lertvittayakumjorn2021explanation} has been released. The survey targets explanatory debugging and human-in-the-loop debugging systems. 

While both papers, \cite{danilevsky2020survey} and \cite{lertvittayakumjorn2021explanation}, targeted a similar framework; there are clear differences between the scopes, depth, and breadth. 
Our work is the first to survey existing work on the explainability of word embeddings and the attention mechanism. Moreover, \cite{danilevsky2020survey} focuses on the interpretability methods that explain individual predictions and \cite{lertvittayakumjorn2021explanation} focuses on interactive human-in-the-loop learning whereas our work is broader. We instead report on literature that interprets the knowledge encoded by language models. Finally, \cite{danilevsky2020survey} claimed that only four papers are found in the literature targeting global explanations which are defined as the study of the predictive process independently of any particular input. While we acknowledge the difficulty of finding such papers due to proper tagging with ExAI keywords, we dedicate three sections (\ref{sec:NLPembeddings}, \ref{sec:RNNs} and \ref{sec:transformers}) to survey over 50 papers, that are missed by \cite{danilevsky2020survey} and we highlight attempts to understand how NLP models process inputs and the information they encode. 

\begin{figure}
    \centering
    \includegraphics[width=0.\textwidth]{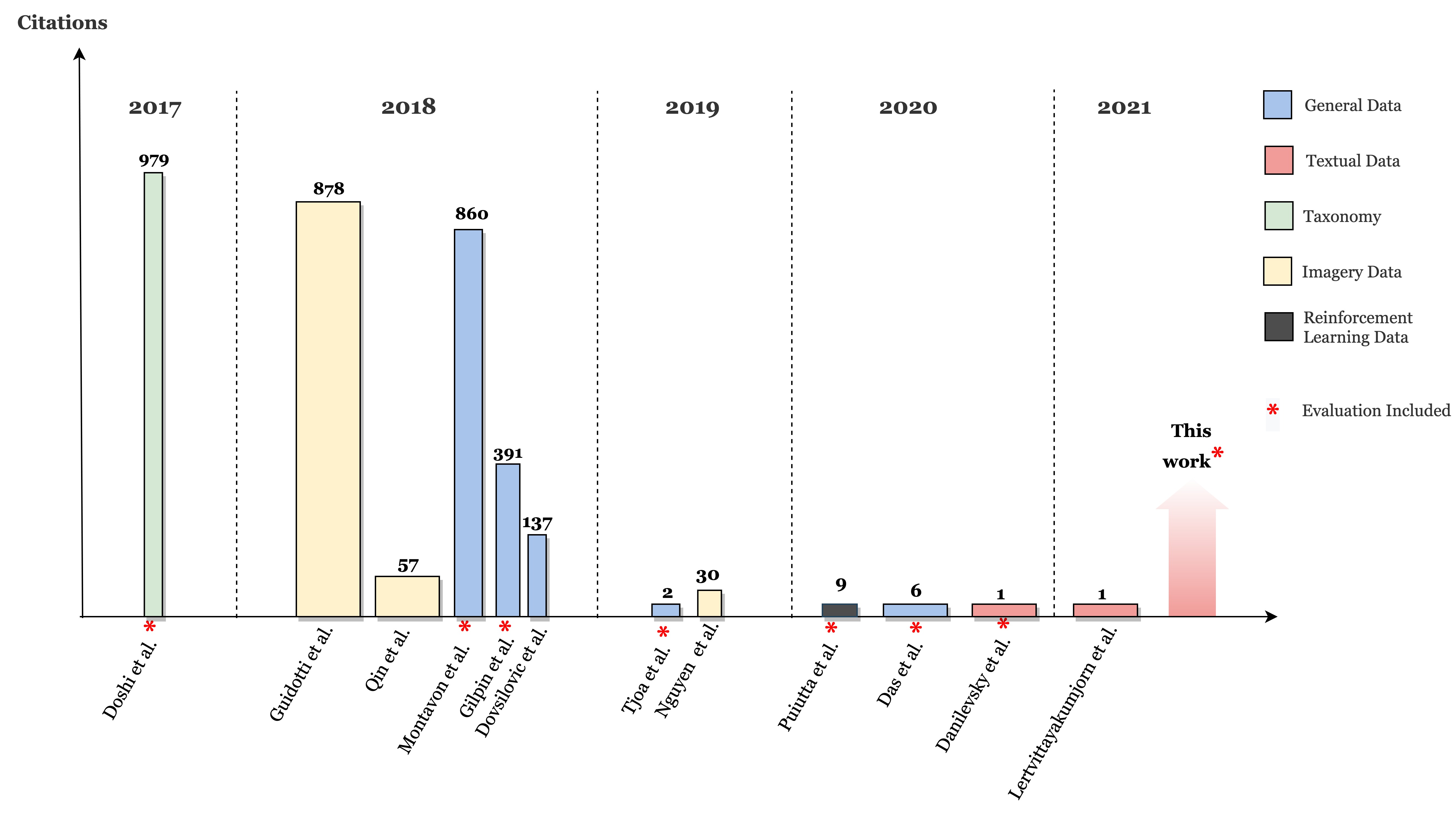}
     \caption{Summary of surveys on ExAI aggregated by year, citation number, evaluation and type of considered data. The thickness of the bar reflects the scope of the presented work categorized into two groups: \textit{broad} and \textit{narrow}. The thick bars represent broad scopes and the thin bars the narrower ones .}
    \label{fig:surveys}
\end{figure}

Figure~\ref{fig:surveys} visualizes existing surveys since 2017 and highlights the broadness of their scope, the data types that they study, their impact on the ExAI community (reflected by their citation number), and whether evaluation methods are included.


%% file: terminology.tex
\begin{figure}
    \centering
    \includegraphics[width=0.9\textwidth]{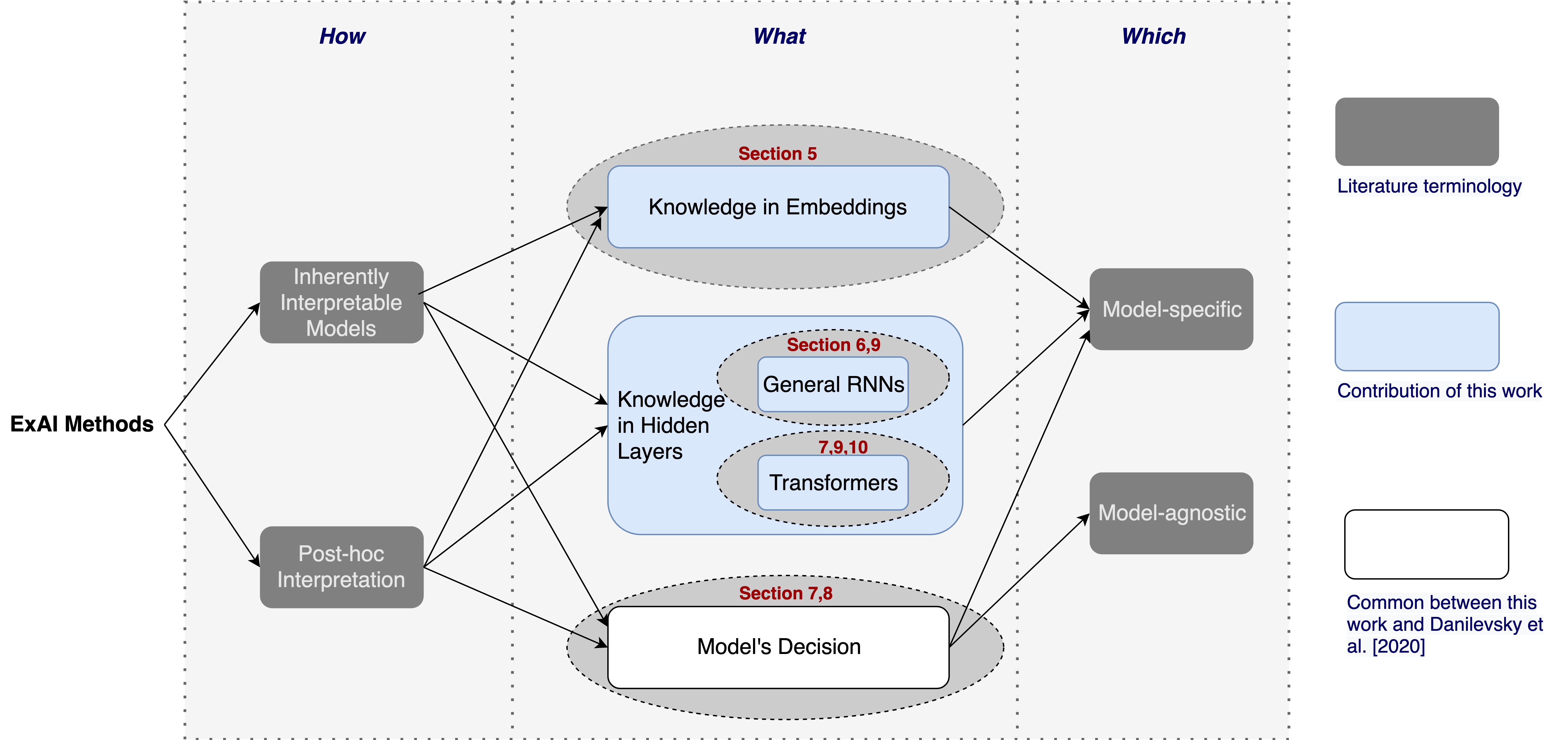}
    \caption{The dimensions studied in this work within terminology used in the ExAI framework and a visual comparison to the work of \cite{danilevsky2020survey}. Missing arrows between consecutive dimensions reflect the infeasibility of the integration of the proposed dimensions.}
    \label{fig:diagram_sections}
\end{figure}

In this survey, we study interpretability methods on NLP models over the  the \textit{how}, the \textit{what}, and the \textit{which} dimensions. 
Our first dimension is specific to \textit{how} the models are being explained, i.e. by design or in a \textit{post-hoc} manner. The former methods develop models from scratch, such that they are \textit{inherently} interpretable while the latter methods explain pretrained black-box models. Although interpretable models are \textit{inherently} transparent, they require modifying the model’s architecture and retraining huge models. A recent study \cite{rudin2019stop}, outlines the difference between both frameworks and highlights the areas where developing explainability methods for black-box models should be avoided in high-stake decision-making environments. 
In the second dimension, we categorize the interpretability attempts into three categories according to \textit{what} they explain. The first category interprets the word embeddings which is on the input level of most of the deep NLP networks. The second category interprets the inner representations of RNNs and transformers which is on the processing level of the DL. Finally, the third category interprets individual model decisions with respect to specific input features or neuron activations that are on the output level of the networks. 
The third dimension addresses the question of \textit{which} models are being interpreted by making the distinction between explanation methods that are model-agnostic, i.e. can operate on any machine learning model, and those that are model-specific, i.e. tailored for specific architectures. Figure~\ref{fig:diagram_sections} highlights the contributions of this paper in terms of these three dimensions within the terminology used in the ExAI framework and highlights the difference between our work and \cite{danilevsky2020survey}.

%% file: nlp_wordembeddings.tex
In this section, we discuss the explainability of word embeddings which constitute the input to the majority of NLP models. Word embedding models are the result of various optimization methods and deep structures that represent words as low-dimensional dense continuous vectors \cite{mikolov2013efficient,pennington2014glove}. Although effective at encoding semantic and syntactic information, those vectors are notoriously hard to interpret. Additionally, they show little resemblance to the theories of lexical semantics, which are based on familiar, discrete classes and relations. 

Interpretability of word embeddings is not only essential for developing transparent models but it is desirable for lexicon induction, efficiency, and fairness \cite{dufter2019analytical}. For instance, if interpretable, semantically or syntactically similar words can be easily extracted in lexicon induction. Additionally, the evaluation embeddings would be transparent by examining the information encoded by the individual vectors. Interpretability has also a computational advantage on many classification tasks where irrelevant dimensions can be disregarded. Finally, \textit{fairness} can be ensured by removing dimensions encoding protected attributes such as gender or race \cite{bolukbasi2016man}. 

Recently, different approaches have been suggested to improve the interpretability of word embeddings. Such methods, discussed next, either rely on visualization \cite{maaten2008visualizing}, impose sparsity \cite{murphy2012learning,faruqui2015sparse,sun2016sparse} and dimension-specific constraints \cite{zhao2018learning}, use rotation techniques \cite{rothe2016word,park2017rotated,ethayarajh2019rotate} or incorporate external knowledge \cite{faruqui2014retrofitting,jha2018interpretable} and contextual information \cite{peters2018deep,peters2018dissecting,tenney2019bert} to derive more interpretable embeddings. Other approaches rely on neighborhood analysis to quantify interpretable characteristics of a given embedding model\cite{rogers2018s}. Such analysis allows to dissect the information that a given word embedding encodes and to explain the correlation between the embedding model on different tasks. All these approaches are \textit{model-specific} and the distinction between \textit{post-hoc} and \textit{inherent} interpretability is later highlighted in Table~\ref{tbl:em_summary}.

\subsection{Sparsification of Embedding Spaces}\label{sec:em_sparse}
\input{nlp_em_sparse}

\subsection{Rotation of Embedding Spaces}
\input{nlp_em_rotation}

\subsection{Integrating External Knowledge}
\input{nlp_em_external}

\subsection{Contextualized Embeddings}
\input{nlp_em_context}

\subsection{Evaluating Embeddings Interpretability}
\input{nlp_em_eval}

\subsection{Discussion}
\input{nlp_em_discussion}

%% file: nlp_em_sparse.tex
\begin{figure}[h]
    \centering
    \includegraphics[width=0.7\textwidth]{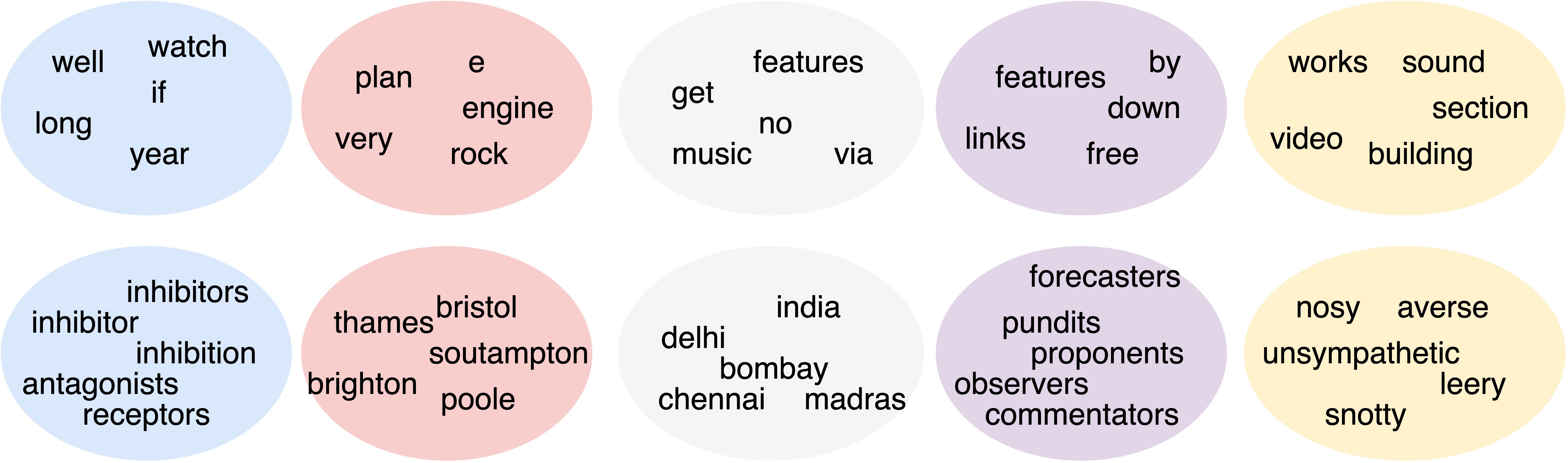}
    \caption{Sets of 5 words for 5 randomly chosen dimensions from a dense (SVD300) and sparse (NNSE1000) embdding model as in \cite{murphy2012learning}. Better semantic coherence is observed in the sparse NNSE1000 model.}
    \label{fig:em_sparse}
\end{figure}

Dense word embeddings cannot solely provide meaningful representations.
For instance, in dense word embeddings, the representation of a word $w$ is a dense vector of small positive or negative values spreading several hundred dimensions. Those dimensions are also active (i.e. having non-zero values) for words of different types and domains. Consequently, an active dimension in a dense word embedding model cannot imply a specific semantic or syntactic form. Sparsifying word embedding models can thus map dimensions to meaning or syntax making embeddings more explainable. To illustrate this concept, we visualize in Figure~\ref{fig:em_sparse}, 5 words for 5 randomly chosen dimensions from a dense (SVD300) and sparse (NNSE1000) embedding models as derived by \cite{murphy2012learning}. We observe that the words in each dimension are more semantically coherent for the sparse embedding model and thereby more interpretable.

Murphy et al. \cite{murphy2012learning} used a matrix factorization algorithm, Non-Negative Sparse Encoding (NNSE), to derive the first distributional model that satisfies sparsity, effectiveness and interpretability. The authors decompose the input representation $\bm{X}\in\mathbb{R}^{m\times n}$ into two matrices, $\bm{A}\in\mathbb{R}^{m\times k}$ and $\bm{D}\in\mathbb{R}^{k\times n}$ subject to sparsity and non-negativity constraints. Their problem is to find $\bm{A},\bm{D}$ that minimize$ \sum_{i=1}^{m} ||\bm{X}_{i,:} - \bm{A}_{i,:}\times \bm{D}||^2 + \lambda ||\bm{A}_{i,:}||_1$, subject to $\bm{A}_{i,j} \ge 0$ and $\bm{D}_{i,:}\bm{D}_{i,:}^\mathsf{T} \le 1$. 

The optimization in the NNSE method requires heavy memory usage and cannot effectively deal with streaming text data. Consequently, an online approach to deriving an interpretable word embedding model (OIWE) is proposed in \cite{luo2015online}. OIWE uses projected gradient descent to apply non-negative constraints on non-negative methods such as Skip-Gram. 
An unconstrained optimization problem is later proposed in \cite{faruqui2015sparse} to transform any distributed representation into sparse or binary vectors. Using sparse coding, each input vector $\bm{X}_i$ is represented as a sparse linear combination of basis vectors. 
Using online adaptive gradient descent, longer and sparser vectors are derived as more interpretable ``overcomplete''
representations.

Later, in \cite{subramanian2018spine}, the authors employed de-noising $k$-sparse auto-encoder to obtain an interpretable transformation of input embeddings, SParse Interpretable Neural Embeddings (SPINE). Input embeddings are projected into a new space $\in \mathbb{R}^{m\times v}$ where embeddings are both sparse and non-negative. De-noising $k$-sparse auto-encoder is used to train the model by minimizing the combination of the reconstruction loss, the average sparsity loss, and the partial sparsity loss over the data set while capturing the sparsity constraints. 

Deriving the embedding models, discussed so far, relies on optimization methods that mainly operate in $\epsilon$-accurate regimes. Operating in a \textit{post-hoc} manner might cause an accumulation of inaccuracies. Some methods introduce sparsity as a post-processing step. For instance, Sun et al. \cite{sun2016sparse} directly apply the sparsity constraint while computing the word embeddings of \textit{Word2Vec}. This is achieved by introducing the \textit{l}$_1$ regularizer and employing regularized dual averaging to produce sparse representations. Additionally, Panigrahi et al. \cite{panigrahi2019word2sense} developed an unsupervised method to generate \textit{Word2Sense} where each dimension encodes a fine-grained \textit{sense}. \textit{Word2Sense} embeddings are probability distributions over \textit{senses} where the probabilities of each word and sense are estimated and the Jensen Shannon divergence \cite{fuglede2004jensen} is then used to compute word similarities to eliminate redundant senses.

%% file: nlp_em_rotation.tex
From a linear algebra perspective, having a \textit{transparent} basis, every embedding vector can be explained as a combination of understandable concepts. To identify these interpretable dimensions (i.e. basis), word spaces can be \textit{rotated} while preserving the encoded information.

Rothe et al. \cite{rothe2016word} formulated this rotation as a decomposition of an embedding space into two components: an interpretable orthogonal subspace and a ``remainder'' subspace. The resulting sub-spaces and their orthogonal complements form the basis for an embedding \textit{calculus} that supports certain operations. The goal becomes to find an orthogonal matrix that transforms the embedding space into an interpretable one with fewer dimensions. Different lexicons such as opinion, POS, and emotion are used to train the transformation matrix to minimize the distance between lexicon pairs and words with identical labels. Consequently, the authors extended the ``king - man + woman=queen'' analogy into operations like ``-1$\times$ hate = love''. 

In \cite{park2017rotated}, interpretability is induced by factor rotation that reforms the word embedding matrix to have a simple structure by a linear transformation. The rotation encourages each row and column (word vector and dimension, respectively) to have a few large values. More specifically, the rotation is a post-processing step that computes a rotated matrix minimizing the rotation criterion. The latter was introduced in \cite{crawford1970general} by forcing a low complexity on the rows and the columns of the rotated matrix and is minimized using the gradient projection algorithm. This approach can be seen as a combination of \textit{sparsification} and \textit{rotation} where embedding models are rotated while encouraging low complexities in the values.
Likewise, \cite{ethayarajh2019rotate} computes orthogonal transformations by randomly sampling word vectors and their target before deriving the orthogonal Procrustes closed-form solution.  

The approaches discussed above enhance the explainability of existing model embeddings. However, the derived explanations are simple linear algebraic analogies, i.e. at the level of simple grammatical relations such as negation, grammatical gender, and prefix derivation. In \cite{allen2019analogies}, the explainability of embeddings was addressed from a different angle. Instead of making the embedding models more explainable, Allen et al. \cite{allen2019analogies} targeted well-known linear relationships and derived their probabilistically-grounded interpretations. Specifically, the authors aim at explaining why the embeddings often satisfy $w_b* = w_a* - w_a + w_b$ when word embeddings are trained using only word co-occurrence.
This is achieved by first defining \textit{paraphrasing} as ``an equivalence drawn between words and word sets by reference to the distributions they induce over words around them''. Then, the authors show that \textit{paraphrasing} determines linear relationships hold whenever the embeddings factorize point-wise mutual information. Finally, the linear relationship follows when analogies between words are interpreted as word transformations sharing the parameters. The mathematical definitions and proofs provided in the paper establish the first rigorous explanation of the embeddings' linear relationships.

%% file: nlp_em_external.tex
The previous two methods discussed the interpretability of word embeddings without the incorporation of external knowledge. However, existing lexicons and ontologies help in building embeddings spaces that can, by construction, reflect the semantic and syntactic relationships and that are thus more interpretable. For instance, Faruqui et al. \cite{faruqui2014retrofitting} enhance the interpretability of embeddings by fine-tuning the embedding model using relational information from semantic lexicons. This is achieved by proposing a graph-based learning framework, ``retrofitting'', for incorporating lexical relational resources. Given an un-directed graph $\Omega=(V,E)$ of semantic relationships and a word embedding model $\bm{\hat{Q}}=(\hat{q}_1, \dots, \hat{q}_n)$, the goal is to learn a \textit{refined} embedding model, $\bm{Q} = (q_1, \dots, q_n)$, such that its column vectors (word embeddings) are close in distance to their counterparts in $\bm{\hat{Q}}$ and to their adjacents under $\Omega$. The optimization problem is thus to minimize $ \sum_{i=1}^{n} \big[ \alpha_i||q_i - \hat{q}_i ||^2 +\sum_{(i,j)\in E}\beta_{ij}||q_i-q_j||^2\big]$
The embeddings are first trained independently of the semantic lexicons and then through ``retrofitting''. $\bm{Q}$ is computed by solving a system of linear equations using an efficient iterative updating method.

In some particular domains, knowledge graphs go beyond syntax and semantics and develop knowledge basis on higher-level categorical and scientific relations. The medical domain is a perfect illustration where \cite{jha2018interpretable}, for instance, incorporates the rich categorical and taxonomic knowledge in the biomedical domain to leverage the interpretability of medical embeddings. The authors learn a transformation matrix that transforms the word embeddings to a new interpretable space according to the biomedical taxonomy while retaining its expressive features. Similarly, Pelevina et al. \cite{pelevina2016making} utilize an \textit{ego-network} to transform word embeddings into sense vectors. The authors define an \textit{ego-network} as a set single nodes, \textit{ego}, along with the nodes they are connected to, \textit{alter}. After learning word embeddings and building a graph of nearest neighbors based on vector similarities, word senses are induced by clustering the\textit{ ego-network}. Such word senses can be effectively used in word sense disambiguation.

Although based on a sound mathematical formulation, the previously discussed approaches do not explicitly show the practicality of the derived explainable embeddings in tasks such as \textit{de-biasing} the embedding models. Instead,  \cite{zhao2018learning} addressed the gender-neutral word embeddings problem and aimed at eliminating gender influence while preserving essential gender information in specific dimensions. This was done by categorizing all the words into male-definition, female-definition, and gender-neutral according to WordNet definitions. Then, embeddings were learned by minimizing using stochastic gradient descent a combination of (1) the word proximities  (2) the negative distance between words in the female and male-definition seed words, and (3) the difference between female words and their male counterpart.

%% file: nlp_em_context.tex
The word embeddings that we discussed so far are functions where every word, regardless of whether it has more than one meaning, has a unique embedding vector. Peters et al. \cite{peters2018deep}, proposed an embedding model, where words are represented by contextualized vectors that model the syntactic and semantic characteristics of a word as well as disambiguation across linguistic contexts, i.e., in the case of \textit{polysemy}. Disambiguation is possible due to the formulation of the word representation as a function of the entire sentence in pre-trained bidirectional language models. While such models are discussed in Section~\ref{sec:transformers}, we focus here on the explainability of their embedding layer rather than the full model. 

The work of \cite{peters2018dissecting} empirically ``dissects'' the contextualized embeddings by evaluating their performance in a suite of four NLP tasks. The authors investigate the intrinsic properties of contextual vectors that are independent of the NLP model and architecture details by studying how semantic and syntactic information is modeled throughout the network's depth. The results show that morphology is encoded in the word embedding layer, the local syntax in the early contextual layers, and semantic information in the deeper layers. 

Tenney et al. \cite{tenney2019you} studied the way sentence structure is modeled across a range of syntactic, semantic, local, and long-range phenomena in contextualized embeddings. Through edge probing, the authors showed that such embeddings outperform their non-contextualized counterparts on syntactic tasks and not on semantic tasks, which shows that they are better at encoding syntax than higher-level semantics. Moreover, contextualized representations are able to encode long-term linguistic information which helps disambiguate longer-range syntactic dependencies relations and structures. Contextualized embeddings are also studied within the gender bias framework in \cite{zhao2019gender}. The analysis showed that some training data contains significantly more male than female entities which gets reflected in embeddings that systematically encode gender information in an unequal manner.

%% file: nlp_em_eval.tex
The research work described above attempts at improving the interpretability of word embeddings. The question that can be immediately asked targets the evaluation of the interpretability of the obtained embedding spaces. While a common rigorous evaluation method, that is based on well-designed bench-marking datasets, is yet to be developed, researchers are currently utilizing a set of interesting experiments summarized in this section.

The interpretability of the sparsification methods is commonly evaluated through word intrusion which seeks to quantify how coherent the dimensions of a learned word representation are. The experiment goes as follows: from the learned representation, a dimension $i$ is chosen and the vocabulary words are then ranked according to the variance of their values. The four highest-ranked words are then chosen across each dimension and a word from the bottom half of the list is then added as an ``intruder'' that human judges are asked to identify. The precision of the human judges across different state-of-the-art sparse embeddings is reported in Table~\ref{tbl:word_intrusion}. In \cite{sun2016sparse}, the word intrusion task is extended and a new evaluation metric that does not require any human assessment. Given that the intruder word should be different from the top four words while those latter words should be similar, the ratio of the distance between the intruder word and top words to the distance between the top words is thus used as a quantitative metric. The higher the ratio is, the more interpretable the embedding state is. 

\begin{table}[h]
\small
\begin{tabular}{llll}
NNSE \cite{murphy2012learning} & SPOWV \cite{faruqui2015sparse} & SPINE \cite{subramanian2018spine} & Word2Sense \cite{panigrahi2019word2sense} \\
\hline\hline
92.33                          & 41.75                          & 74.83                             & 75.3  \\    
\end{tabular}
\caption{Precision on the word intrusion task on the state-of-the-art sparse embedding spaces}\label{tbl:word_intrusion}
\end{table}

Additionally, Rothe et al. \cite{rothe2016word} used the cosine similarity to decide whether two words are synonyms or antonyms which reflects the level of interpretability of word connotations. The cosine similarity between embeddings was also used in polarity spectrum creation where the main task is to predict the spectrum of a word in a certain query. The morphological analogy was also used to test how well these embeddings can encode the POS tags in the computed subspaces. These experiments test the quality of a subspace rather than its interpretability. 

The work in \cite{faruqui2014retrofitting} evaluates the interpretability of embeddings as their ability in capturing semantic and syntactic aspects. Such tasks include the word similarity and the synonym-selection tests which are aligned with \cite{rothe2016word} discussed above. The syntactic relations test evaluates how well the embeddings can encode relations of the form ``\textit{a} is to \textit{b} as \textit{c} is to \textit{d}''. Sentiment analysis tests the knowledge encoded in these representations. While these tests better target the interpretability of the representations in terms of their semantic and syntactic relations, they do not explicitly reflect the interpretability of certain dimensions.

Peters et al. \cite{peters2018deep} dissect the word representations in bi-directional models and evaluate their interpretability in terms of the quality of the interpretation on tasks such as semantic role labeling, and constituency parsing, and named entity recognition. Later, this set of experiments is extended in \cite{peters2018dissecting} to include question answering, sentiment analysis, and textual entailment. 

%% file: nlp_em_discussion.tex
The evaluation methods discussed above mainly reflect the quality of embeddings rather than their interpretability. For instance, evaluating the embeddings on a POS tagging task shows how well they encode some syntax properties without reflecting the exact dimension, or the set of dimensions, where such properties are encoded. In other words, the majority of experiments report an overall performance on some NLP tasks without addressing the alignment between a dimension and a particular context or syntactic aspect.

Some interesting experiments are reported in \cite{murphy2012learning,faruqui2015sparse,subramanian2018spine,panigrahi2019word2sense}. Murphy et al. \cite{murphy2012learning} qualitatively assess the interpretability of dimensions of the NNSE embeddings by investigating the dominating dimensions in the NNSE representations. 
Similarly, Faruqui et al. \cite{faruqui2015sparse} consider the dimension interpretability and qualitatively assessed whether the top-ranking words for a particular dimension display semantic or syntactic groupings. 
\cite{subramanian2018spine} extend this experiment to SPINE and qualitatively compare the interpretability of SPINE to that of SPOW with two baseline embedding models, \textit{word2vec} and \textit{Glove}. The authors examine the top participating dimensions for some sampled words and study the top words from the participating dimensions. Similar experiments are conducted by \cite{panigrahi2019word2sense} in \textit{word2sense}.

\begin{figure}[h]
    \centering
    \includegraphics[width=0.4\textwidth]{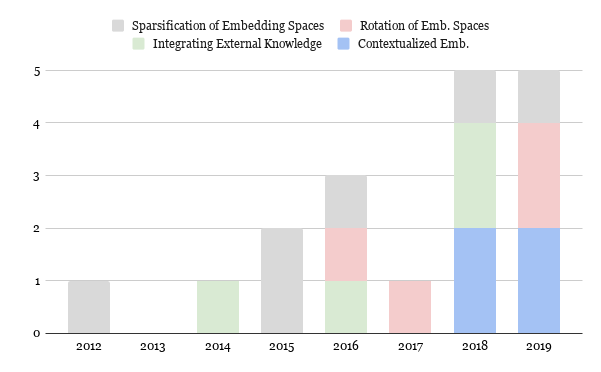}
    \caption{Number of the surveyed papers in each method reflecting the interest since 2012.}
    \label{fig:em_hist}
\end{figure}

These experiments are paving the way towards a rigorous evaluation of embeddings' explainability but are still lacking the commonality and objectivity aspects. Finally, the majority of these methods operate on embedding models and are thus \textit{model-specific}. However, some of the obtained models are \textit{inherently} interpretable; whereas others need further processing. Table~\ref{tbl:em_summary} summarizes the discussed methods within the dimensions discussed in this work while highlighting their reliance on common evaluation schemes and whether they need existing embedding spaces. Figure~\ref{fig:em_hist} reflects the interest in the interpretability of word embeddings by showing the number of papers we surveyed within each category since 2012. One can see that the sparsification of embeddings was one of the earliest methods in the field. After the success of bidirectional transformer models in 2017, contextualized embeddings are attracting researchers to study their interpretability. Interest in other methods seems to be consistent and a general interest in the interpretability of word embeddings seems to be increasing.

\begin{table}[h]
\small
\begin{tabular}{p{5cm}cccl}

\textbf{Method} &
\textbf{Needs Existing} &
\textbf{Common} &
\textit{\textbf{Posthoc}} / &
\textbf{References} \\
 &
\textbf{Embedding}? &
\textbf{Evaluation} &
\textbf{\textit{Inherently}} &
 \\
  
 \hline
Sparsification &
  \checkmark &
  \checkmark &
  \textit{Inherently} &
  \cite{murphy2012learning,luo2015online,faruqui2015sparse} \cite{subramanian2018spine} \\
 
   &
  x &
  \checkmark &
  \textit{Inherently} &
  \cite{sun2016sparse,panigrahi2019word2sense} \\
  \hline\\
Rotation &
  \checkmark &
  x &
  \textit{Inherently} &
  \cite{rothe2016word,allen2019analogies,park2017rotated,ethayarajh2019rotate} \\
  \hline
Integrating External Knowledge &
  \checkmark &
  x &
  \textit{Inherently} &
  \cite{faruqui2014retrofitting,jha2018interpretable,pelevina2016making,zhao2018learning} \\
  \hline
Contextualized Embeddings &
  \checkmark &
  x &
  \textit{Post-hoc} &
  \cite{peters2018dissecting,tenney2019you,zhao2019gender} \\

   &
   &
   &
  \textit{Inherently} &
  \cite{peters2018deep}\\
  \hline
\end{tabular}
\caption{Summary of existing work on the interpretability of word embeddings}\label{tbl:em_summary}
\end{table}

%% file: rnn.tex
After discussing the interpretability on the input level, we focus next on the processing level. Hence, we dissect the inner representations of  RNNs. These methods either implement \textit{inherently} interpretable RNN architectures or try to interpret existing RNN architectures in a \textit{post-hoc} manner. Both approaches however operate in a \textit{model-specific} way.

\subsection{\textit{Post-hoc} Interpretation}
Most of the \textit{post-hoc} interpretation methods try to dissect hidden knowledge in trained deep networks from syntax and semantic lens. One such lens is \textit{compositionality}, defined in \cite{li2016visualizing}, as a way to understand how such networks build sentence semantics from individual words. The authors develop strategies based on heatmap and t-sne visualization by plotting unit values to visualize negation, intensification, and concessive clauses. Similar to \cite{kadar2017representation}, but using the unit's salience, the authors are able to compute the amount that each input contributes to the semantics of a sentence. These strategies show that Long Short-Term Memory Networks (LSTMs) are able to perform well due to their ability to maintain a sharp focus on essential keywords. Moreover, sentiment analysis tasks are sensitive to some dimensions in an embedding. More importantly, neural models in NLP are able to learn the properties of local \textit{compositionality} in a subtle way while respecting asymmetry as in negation. 

Focusing on syntax, Blevins et al. \cite{blevins2018deep} develop a set of experiments that investigate how the internal representations in deep RNNs capture soft, hierarchical syntactic information without explicit supervision. First, the authors extract word-level representations produced by layers of the RNN trained on tasks such as dependency parsing,  machine translation, and language modeling. A classifier is then trained to predict the POS tag and the parent, grand-parent, and great-grandparent constituent labels of that word. Another classifier is trained to check whether a dependency arc exists between two words. The evaluation of these classifiers shows that the representations learned by RNNs encode syntax beyond the explicit information encountered during training. The results demonstrate that the word representations produced in RNNs at different depths are highly correlated with syntactic features. In particular, deeper layers are shown to capture higher-level syntax notions. 

These findings agree with \cite{tran2018importance}, where recurrent architectures are compared to non-recurrent ones for their ability to model hierarchical syntax. Experiments on subject-verb agreement and logical inference show that recurrent architectures, LSTM specifically, are notably more robust and present better generalization guarantees when longer sequences are encountered. The same quest is applied to LSTMs in \cite{linzen2016assessing} to study how syntactic structures are encoded. Through number agreement in English subject-verb dependencies experiments, the authors show that LSTMs are able to capture a considerable amount of grammatical structures, but more expressive architectures may be required to reduce errors in particularly complex sentences. Number agreement is also studied later by \cite{gulordava2018colorless} by including nonsensical sentences where RNNs cannot rely on semantic or lexical cues and by comparing to human intuition to show that long-distance agreements are reliably encoded in an RNN. In \cite{khandelwal2018sharp}, the role of context in LSTMs is analyzed through ablation studies by shuffling, replacing, and dropping prior context words. LSTMs were found to be able to use about 200 tokens of context on average. However, LSTMs struggle to distinguish between nearby context from distant history. More interestingly, the model pays attention to the order of words only in the recently-processed sentence. LSTMs' predictions have been empirically explored in \cite{karpathy2015visualizing} along with the LSTMs' learned representations. Specifically, long-range dependencies are investigated in character-level language to seek cells that identify high-level patterns such as line lengths, brackets, and quotes. Karpathy et al. \cite{karpathy2015visualizing} study $n$-gram model to conclude that LSTMs' performance improves on characters that require long-range reasoning.

The discussed approaches address particular semantic and syntactic structures and study \textit{where} and \textit{how} these structures are encoded. Nonetheless, these approaches do not study how the input is processed through these inner encodings. Very recently, Hou et al. \cite{hou2020learning} suggested a novel interpretation framework inspired by computation theory. The authors are the first to draw the analogy between Finite State Automata (FSA) and an interpretable inner mechanism for sequential data processing. In their work, FSAs are learned from trained RNN models to achieve better interpretability of the latter. When inputting several sequences to an RNN model, the hidden state points are gathered, and similar ones are clustered. The FSA states are the clusters, and the transitions between the states arise when one item of the input sequence is processed. Aside from the novel connection with FSA, an interesting advantage of \cite{hou2020learning} is its ability to provide global explanations in classification tasks. 

\subsection{\textit{Inherently} Interpretable Models}
\textit{Inherently} interpretable RNNs are trained in an explainable way by adding transparency constraints \cite{wisdom2016interpretable} and exploring tree and graph-like structures \cite{tai2015improved,liang2017interpretable}. In \cite{wisdom2016interpretable}, the authors develop an \textit{inherently} interpretable RNN, SISTA-RNN, based on the sequential iterative soft-thresholding algorithm and the idea of \textit{deep unfolding }\cite{hershey2014deep}. By allowing the RNN parameters to be seen as the parameters of a probabilistic model, the weights and outputs can retain their meaning. Traditionally, given input-output training pairs $(\bm{x}_i, \bm{y}_i)_{i=1,\dots,N}$ and model parameters $\bm{\theta}$, RNNs learn $\bm{\theta}$ according to the following optimization problem: $\underset{\bm{\theta}}{\min} \sum_i f(\bm{y}_i, \bm{\hat{y}}_i)$ subject to $  \bm{\hat{y}}_i = g_{\theta}(\bm{x}_i), i=1,\dots,N $, 
where $f$ is the loss function and $g$ is the conventional black-box RNN. By changing the optimization constraint to $\bm{\hat{y}}_i = h_{\theta}(\bm{x}_i), i=1,\dots,N$ and $h_{\bm{\theta}}(\bm{x}_i) = \underset{z}{\arg\min}\mathcal{P}_{\theta}(\bm{z},\bm{x}_i)$ with $h$ attempting to solve an optimization problem $\mathcal{P}_{\bm{\theta}}$ that corresponds to a principled probabilistic model, the parameters $\bm{\theta}$ become interpretable. SISTA-RNN is proved to achieve a higher performance on a sequential compressive sensing task.

Tai et al. \cite{tai2015improved} improved the semantic representation of LSTMs by building a semantic tree on semantic topologies. Tree-LSTMs are similar to the traditional LSTM units but differ in the gating vector and memory cell update that are dependent on the state of the children (tree notion). A forget gate per child allows Tree-LSTM to select the information in each child.
Later, Liang et al. \cite{liang2017interpretable} learn interpretable representation following a wise graph construction paradigm. Their \textit{structure-evolving} LSTM first considers each data element as a separate node in a graph. Then, nodes with high compatibility are recursively merged to form richer and more interpretable encodings. Similarly, graph LSTMs \cite{peng2017cross} are formulated to incorporate linguistic analysis. This approach is shown to encode richer linguistic knowledge improving the performance of relation extraction \cite{peng2017cross,guo2019attention,zhou2020graph,zhang2018chinese}. Although both LSTMs outperformed existing systems, their evaluation remains under-investigated. For instance, \cite{tai2015improved} evaluate their LSTMs on sentiment analysis and semantic relatedness tasks. Moreover, the effectiveness of   \textit{structure-evolving} LSTMs is mainly evaluated on a non-NLP-related image segmentation task without thorough testing on other modalities.

Neural module networks jointly train deep ``modules'' for visual question answering offering visual explanations for an answer \cite{andreas2016neural,hu2017learning}. While such networks are not fully within a textual modality, i.e. they accept the question as text and an image as a reference, we briefly discuss them in this survey for completeness purposes. Neural module networks learn to parse questions as executable modules to understand well synthetic visual QA domains \cite{gupta2019neural}. The questions are analyzed in a semantic parser before determining the basic computational units that contribute to the answer \cite{andreas2016neural}. Andreas et al. \cite{andreas2016learning} learn the parameters for these modules jointly via reinforcement learning on (world, question, answer) triples. Gupta et al. \cite{gupta2019neural} extend this approach and introduce modules that reason over a paragraph of text through symbolic reasoning over numbers and dates. Furthermore, an unsupervised auxiliary loss is suggested to extract arguments explaining specific events in the text. 

\input{rnn_discussion}

%% file: rnn_discussion.tex
\subsection{Discussion}
Table~\ref{tbl:rnns} summarizes the aforementioned approaches while highlighting their interpretability and evaluation methods. While \cite{li2016visualizing} is inspired by a non-text-specific interpretability approach, the rest of the methods are designed to suit NLP models. One can clearly see that a common evaluation is not established yet, some methods can be subjective \cite{gulordava2018colorless} while others rely on human-annotated data \cite{blevins2018deep,tai2015improved}. Such annotations are not directly related to explainability; they however reflect some semantic or syntactic aspects such as subject-verb agreement or POS tags that. Although a good performance can imply that the specific semantic or syntactic aspect is indeed encoded in the model, it cannot directly imply explainability. Localization of the encoded aspect in the model as well as input analysis in terms of these aspects are further needed to achieve pure model transparency.

\begin{table}[h]
\small
\begin{tabular}{p{2.2cm}p{2cm}p{1.8cm}p{4.5cm}p{4.2cm}}
\hline
\textbf{Reference} &
  \textit{\textbf{\textit{Inherently}/ \textit{Post-hoc}}} &
  \textbf{Model} &
  \textbf{Method} &
  \textbf{Evaluation Method} \\
  \hline
  \hline
  
  \cite{karpathy2015visualizing} (2015) &
  \textit{Post-hoc} &
  LSTM &
  Identification of high-level patterns & Reasoning on $n-$gram models, Subject-verb agreement \\
  
\cite{li2016visualizing} (2016) &
  \textit{Post-hoc} &
  LSTM &
  Heatmap &
  T-SNE visualization \\

\cite{linzen2016assessing} (2016) &
  \textit{Post-hoc} &
  LSTM &
  Probing &
  Subject-verb agreement and logical inference \\
  \cite{blevins2018deep} (2018) &
  \textit{Post-hoc} &
  RNN &
  Correspondence between network depth and syntactic depth &
  POS tag and dependency classification \\
\cite{gulordava2018colorless} (2018) &
  \textit{Post-hoc} &
  RNN &
  Comparison to human intuition &
  Subject-verb agreement \\
\cite{khandelwal2018sharp} (2018) &
  \textit{Post-hoc} &
  LSTM &
  Ablation &
  Context localization \\

\cite{khandelwal2018sharp} (2018) &
  \textit{Post-hoc} &
  LSTM &
  Ablation &
  Context localization \\  
  
  \hline
  \cite{tai2015improved} (2015) &
  \textit{Inherently} &
  LSTM &
  Integration of semantic topologies &
  Sentiment analysis and relation classification \\
  
\cite{wisdom2016interpretable} (2016) &
  \textit{Inherently} &
  RNN &
  Sequential iterative soft-thresholding algorithm and deep unfolding &
  Sequential compressive sensing task \\

\cite{liang2017interpretable} (2017) &
  \textit{Inherently} &
  LSTM &
  Integration of knowledge graphs &
  Semantic object parsing\\

  \cite{andreas2016neural} (2016) &
  \textit{Inherently} &
  Module Networks &
  Reinforcement Learning &
  Visual question answering\\

  \cite{hu2017learning} (2017) &
  \textit{Inherently} &
  Module Networks &
  Joint learning &
  Visual question answering\\

  \cite{gupta2019neural} (2019) &
  \textit{Inherently} &
  Module Networks &
  Unsupervised auxiliary loss &
  Reasoning over text\\  
  
  \cite{peng2017cross,guo2019attention,zhou2020graph,zhang2018chinese} (2017-2020) &
  \textit{Inherently} &
  LSTM &
  Encoding of richer linguistic knowledge &
  Relation Extraction\\ 
  
  \hline
\end{tabular}
\caption{Summary of interpretability attempts at understanding the inner representations of RNNs}\label{tbl:rnns}
\end{table}

%% file: transformer.tex
After discussing the interpretability of general RNNs, we consider the popular transformer models \cite{vaswani2017attention}. Such models are based on attention mechanisms to handle ordered sequences of data. Transformers follow the encoder-decoder structure using stacked multi-head self-attention and fully connected layers. Bidirectional Encoder Representations from Transformers (BERT) \cite{devlin2018bert} and GPT-2 \cite{radford2019language}, have been trained on huge text corpora and are currently used as state-of-the-art NLP models after fine-tuning on specific tasks. Different approaches have been studied to understand the inner dynamics of transformer models and to visualize the attention weights in order to better understand \textit{how} they process input and \textit{why} they do it so well. All the discussed approaches are \textit{model-specific} and operate in a \textit{post-hoc} manner. 

\input{trans_vis}

\input{trans_attn}

\input{trans_bert}

\input{trans_discussion}

%% file: trans_vis.tex
\subsection{Visualization of Transformers}
The Explainability of transformers has been extensively addressed from a visualization perspective by developing tools that allow the user to interact with such models to understand their inner mechanisms. For instance, the work of \cite{lee2017interactive} presents an interactive tool to visualize attention and provides an interface to dynamically adjust the search tree and attention weights. Similarly, in \cite{liu2018visual}, the authors proposed a flexible visualization library to visually analyze models for natural language inference and machine comprehension that relies on a perturbation-driven exploration strategy. Also, in \cite{strobelt2018s}, SEQ2SEQ-VIS, visual analytics is presented for sequence-to-sequence model debugging by visualizing the five stages of a seq2seq model: encoder, decoder, attention, prediction, and beam search. SEQ2SEQ-VIS also describes the knowledge that the model has learned by relying on transitions of latent states and their related neighbors. Finally, SEQ2SEQ-VIS provides an interactive method to manipulate the model internally and observe the impact on the output.  

ExBERT \cite{hoover2019exbert} is another interactive visualization tool that uses linguistic annotations, masking, and nearest neighbor search to provide insights into contextual representations learned in transformer models in three main components: (1) the attention, 
(2) the corpus
and (3) the summary view.
Similarly, Vig \cite{vig2019visualizing} presents a tool to visualize BERT and GPT-2 models at three different granularity levels: the attention-head level, the model level, and the neuron level. Table~\ref{tbl:trans_vis} summarizes these methods and the specific models they are applied on.

\begin{table}[h]
\small
\begin{tabular}{lp{6.5cm}p{6cm}}
\hline
\textbf{Reference} & \textbf{Method} & \textbf{Model} \\
\hline\hline
\cite{lee2017interactive} (2017) & Visualization of attention weights & Neural machine translation models \\
\cite{liu2018visual} (2018) & Perturbation-driven & Natural language inference and machine comprehension models \\
\cite{strobelt2018s} (2018) & Visualization of encoder, decoder, attention, prediction, and beam search & Sequence-to-sequence models \\
\cite{hoover2019exbert} (2019) & Linguistic annotations, masking, and nearest neighbor search & BERT \\
\cite{vig2019visualizing} (2019) & Visualization on the level of attention-head, model, neuron & BERT and GPT-2\\
\hline
\end{tabular}
\caption{Summary of transformer visualization methods}\label{tbl:trans_vis}
\end{table}

%% file: trans_attn.tex
\subsection{Is the Attention Mechanism \textit{Inherently} Interpretable?}\label{sec:trans_attn}
Controversy has accompanied attention mechanisms since their introduction. While some attention weights can provide reliable explanations \cite{mullenbach2018explainable}; some researchers showed that attention distributions are not easily interpretable and require further processing  \cite{brunneridentifiability,jain2019attention}. 

In an attempt to investigate these controversies, Vashishth et al. \cite{vashishth2019attention} manually analyzed attention mechanisms on several NLP tasks. The experiments showed that attention weights are interpretable indeed and are correlated with feature importance measures capturing several linguistic notions. Similar findings were obtained by \cite{mullenbach2018explainable} on medical tasks where attention weights were efficient in the selection of most relevant segments in a medical document.

These methods, however, are specific to particular domains and linguistic notions and might not be easily extendable to higher-level knowledge structures. Hence, a parallel line of work attempted at proving that ``attention is not explanation'' by considering more general correlations experiments. For instance, \cite{jain2019attention} show that there is no frequent correlation between the attention weights and feature importance methods. Moreover, the authors identify attention distributions that yield equivalent predictions through correlation computation. 

Prior to these researches,  \cite{rocktaschel2015reasoning} focused on the reasoning capabilities of transformers by considering the recognizing textual entailment task. For this purpose, the attention patterns are visualized on hand-picked validation samples. Word-by-word attention is shown to resolve synonym relationships, match multi-word expressions to single words, and ignore irrelevant parts. Moreover, when a deeper semantics level or common-sense knowledge connects two sentences, attention seems to be capable of capturing the semantics. However, when two sentences are not related, attention seems to be dominated by the last output vector. 

The authors in \cite{vig2019analyzing} study the interaction between the syntax and the attention weights in GPT-2 \cite{radford2019language}. The authors investigate the alignment between syntactic dependency and attention weights. The authors aggregate over the corpus the percentage of total attention of a given head that attends to tokens belonging to the given POS tag with syntactic features. The same aggregation method is used to test the alignment between attention and dependency by computing the proportion of attention between tokens that are components of dependency relations. To explore how long- and short-distance relations are captured, the number of tokens spanned by each head is computed, then attention dispersion is computed based on entropy. The heatmap shows that most POS tags are disproportionately targeted by one or more heads. Attention heads that focus on one POS tags vary according to layer depths: determiners are targeted in the early layers, while proper nouns are targeted in deeper layers.
Additionally, the alignment between attention and dependency relations is strongest in the middle layers. Heads in the early layers tend to focus on position rather than content,  whereas attention heads in deeper layers target specific constructs. Finally, deeper layers capture longer-distance relationships. A moderate correlation is found between the distance and entropy of attention and the attention distance is negatively correlated with dependency alignment. 

A common limitation of the discussed approaches is the lack of a unified definition of the explainability of attention. 
Formal explainability definition in NLP, or the lack thereof, is addressed in \cite{brunneridentifiability}, where \textit{identifiability} of the attention weights is defined as their ability to be uniquely identified from the attention head's output. The study of \textit{identifiability} is done on attention weights and token contextualized embeddings and the aggregation of context into hidden tokens. Input tokens are shown to retain their identity whereas the information identity gradually decreases with depth.
Pruthi et al. \cite{pruthi2020learning} manipulate the attention weights to whitewash problematic tokens in explanations that affect the model fairness or accountability. This is achieved by diminishing the weight assigned to these impermissible tokens. 

Table~\ref{tbl:attn} summarizes the controversy around the attention weights while highlighting the methods and the findings of each work. In summary, attention weights are more \textit{inherently} interpretable than the parameters of general deep networks. However, some aspects of interpretability need further processing and investigation of the attention weights.

\begin{table}[h]
\small
\begin{tabular}{lp{4.5cm}cp{6cm}}
\textbf{Reference} & \textbf{Method} & \textbf{Interpretable?} & \textbf{Finding} \\
\hline\hline
\cite{mullenbach2018explainable} (2018) & Selection of most relevant segments& \checkmark & Ability of attention mechanism to identify meaningful explanations \\
\cite{vashishth2019attention} (2019) & Analysis of weights on text classification and text generation tasks & \checkmark & High correlation between attention weights and feature importance of linguistic features \\
\hline\\
\cite{rocktaschel2015reasoning} (2015)& Weights visualization on recognizing textual entailement & X & Dominance of the last output vector over attention in some cases \\
\cite{ding2017visualizing} (2017)& Layer-wise relevance propagation (LRP) \cite{bach2015pixel} & X & Importance of LRP to further interpret the attention weights and the internal workings of transformers \\
\cite{vig2019analyzing} (2019)& Alignment between syntactic dependency and attention through visualization and aggregation & X & (1) Disproportionality between heads targeting POS, (2) Capturing of longer-distance relationships by deeper layers and (3) Moderate correlation between distance and entropy of attention \\
\cite{jain2019attention} (2019)& Correlation and counterfactuals & X & No frequent correlation between attention weights and gradient-based measures of feature importance\\
\cite{brunneridentifiability} (2019) & Aggregation of context into hidden tokens & X & Preservation of the token \textit{identifiability} throughout the model and decrease of information identifiability with depth \\
\cite{pruthi2020learning} (2020) & Diminishing attention weights of impermissible tokens & X & Attention-based explanations can be deceived especially within the fairness context\\
\hline\\
\end{tabular}
\caption{Summary of the literature discussing if the attention weights are inherently interpretable.}\label{tbl:attn}
\end{table}
\hspace{-2em}

%% file: trans_bert.tex
\subsection{Interpretability of BERT}
Bidirectional training is introduced in BERT \cite{devlin2018bert} revolutionizing the training of language models. This section discusses the interpretability methods that study the \textit{why} and the \textit{how} such training works so well. Such dissection aids in understanding BERT \cite{tenney2019bert,jawahar2019does}, its weights \cite{raganato2018analysis,clark2019does} and limitations \cite{rogers2020primer}. More interestingly, it improves the performance of query retrieving by automatically discovering better prompts to use to retrieve and combine more accurate answers as in \cite{jiang2020can}. The approaches discussed next describe the inner workings of BERT models, their attention weights, and methods used to dissect their inner knowledge.

The work of \cite{raganato2018analysis} can be considered as a general evaluation scheme for BERT attention weights. The authors proposed several methods to analyze the linguistic information in BERT. First, heatmaps of attention weights are explored to find linguistic patterns. Second, a maximum spanning tree is constructed for each sentence to check if the syntactic dependencies between tokens have been learned by the network. Sequence labeling tasks are also used to measure how important the learned features are for different tasks. Finally, the encoder weights of a high-resource language pair are used to initialize a low-resource language pair in order to assess the generality of the learned features. These methods lead to insightful conclusions. Attention is shown to follow four different patterns: paying attention to the word itself, to the adjacent words, and to the end of the sentence. Early layers tend to focus on short dependencies while higher ones focus on long dependencies. As expected, training on larger datasets can induce better syntactic relationships, and syntactic dependencies get to be significantly encoded in at least one attention head in each layer of the model.  Moreover, early layers encode syntactic information, whereas semantic information is encoded in the upper layers, and starting from the third layer, information about the input length starts to vanish. 

A study more focused on the attention heads is presented in \cite{clark2019does}. By means of aggregation, most of the heads are found to put little attention on the current token, whereas there is a considerable amount of heads that heavily attend to adjacent tokens, especially in the early layers. Moreover, the authors speculate that, when the head's attention function is not applicable, the head attends to the end of the sentence token. Thus, gradient-based measures of feature importance are applied to show that attending to some end-of-sentence token does not have a substantial impact on BERT's output. In order to study whether attention heads span a few words or attend broadly over many words, the average entropy of each head's attention distribution is computed. The results show that in the early layers, attention heads have widespread attention. To study the alignment between syntactic dependencies and attention weights, attention maps are extracted from BERT. Evaluation of the direction and the prediction of the head shows that certain attention heads specialize in certain dependency relationships. Finally, the coreference resolution test, where the antecedent selection accuracy is computed, shows that BERT heads achieve reasonable co-reference resolution performance. 
 
While \cite{clark2019does} and \cite{raganato2018analysis} are focused on the attention heads and weights, Tenney et al. \cite{tenney2019bert} develop a layer-based approach to interpret the encoded knowledge in BERT's layers. Tenney et al. study the traditional NLP steps that BERT follows to investigate where linguistic information is formed. The authors employ probing techniques to understand the interactions within BERT better and to study at which layers the BERT network can resolve syntactic and semantic structures. It is worth mentioning that the edge probing technique, introduced in \cite{tenney2019bert}, aims at quantifying the degree to which linguistic structures can be extracted from a pre-trained encoder. 
Two metrics are defined: a scalar mixing weight and a cumulative scoring. The former specifies the most relevant combination of layers when a probing classifier is tested on  BERT, whereas the latter quantifies the score improvement on a probing task when a particular layer is considered. The results show that the traditional NLP steps are implicitly followed in order by the network: POS tags are processed earliest, then constituents, dependencies, semantic roles, and coreference are processed in the deep layers. In accordance with previous work, basic syntactic information is encoded in the early network layers,  while higher-level semantic information appears in deeper layers. Moreover, syntactic information is more localizable in a few layers, whereas semantic information is generally spread across the network. 

Jawhar et al. \cite{jawahar2019does} address the same question as \cite{tenney2019bert} and attempt at unpacking the elements of language structure learned by the layers of BERT. The authors also use probing techniques to show phrasal representation in the lower layers of BERT is responsible for phrase-level information while richer hierarchy can be found in intermediate layers with surface features at the bottom, syntactic features in the middle, and semantic features at the top. Finally, the deepest layers encode long-distance dependency such as subject-verb agreement. Jawhar et al. \cite{jawahar2019does} also show that similar to tree-like structures, the linguistic information is encoded in a classical compositional way.

BERT's ability to encode factual knowledge is studied by \cite{petroni2019language,roberts2020much} through linguistic tasks such as facts about entities, common sense, and general question answering. BERT is shown to encode relational knowledge without any fine-tuning and without access to structured data like other NLP methods. This suggests that language models trained on huge corpora can serve as an alternative to traditional knowledge bases \cite{petroni2019language}. Roberts et al. \cite{roberts2020much} consider finetuning language models (such as BERT) to study their ability to encode and retrieve knowledge through natural language queries. The authors found that language models are able to perform well on factual tasks but are expensive to train and more opaque than shallow methods. Moreover, they don't provide any guarantees that knowledge can be updated or removed over the course of training.

Readers are referred to \cite{rogers2020primer} where Rogers et al. survey existing work that dissects how and where BERT encodes semantic, syntactic, and word knowledge. The dissection is done via self-attention heads and throughout BERT layers. Moreover, \cite{rogers2020primer} surveys modifications to BERT's training objectives and architecture, describe the over-parameterization issue in BERT training, and report some of the compression approaches and future research directions.

%% file: trans_discussion.tex
\subsection{Discussion}
Due to their reliance on the attention mechanism, the interpretation of transformers is less challenging than RNNs. In view of the way they are designed, attention weights are relatively more interpretable than the conventional deep networks weights. This design made the evaluation of transformers' inner workings more feasible by visualizing their weights and hidden representations \cite{lee2017interactive,strobelt2018s,vig2019visualizing}. The controversy that accompanied the attention mechanisms trying to address the extent to which the attention weights are explainable can lead to the following conclusion. Relative to general deep networks weights, attention weights can be thought of as more interpretable. However, solely, their ability to provide full transparency or meaningful explanations is questionable. Further processing is needed to achieve proper transparency of transformer models especially for long-distance temporal relations. More specifically, when the task in hand is not a simple classification but a more complex task such as translation, question answering and natural language inference, attention weights might not offer the desired interpretability \cite{ding2017visualizing,jain2019attention}. Mohankumar et al. \cite{mohankumar2020towards} argue that when the attention distribution is computed on input representations that are very similar to each other, they cannot provide very meaningful explanations. For this purpose, the authors of \cite{mohankumar2020towards} diversify the hidden representations over which the distribution are computed for more faithful explanations. Moreover, attention weights become less interpretable in deep layers \cite{vig2019analyzing,brunneridentifiability}. Hence, if the explanation at deeper layers requires further processing especially that output vectors might dominate the attention weights as in \cite{rocktaschel2015reasoning}.

Moreover, similar to RNNs (Section~\ref{sec:RNNs}), the evaluation of interpretability methods on transformers is achieved by designing methods that test specific semantic or syntactic aspect through alignment \cite{vig2019analyzing}, correlations \cite{jain2019attention} and general explainability methods such as LRP \cite{bach2015pixel} and weight visualization \cite{rocktaschel2015reasoning}. The first building block in the common explainability framework of transformers is presented in \cite{brunneridentifiability}. However, their definition and metric need to be extended to involve other syntactic and semantic explainability aspects discussed in this section.
Finally, developing only \textit{post-hoc} interpretation methods on transformers can be explained by the fact that such models are already famous and retraining them is very computationally heavy. Training BERT model on huge corpora, for instance, requires the same energy that five cars consume on average during their lifetime according to \cite{strubell2019energy}.

%% file: exp_decisions.tex
While the previous section opens the deep \textit{black-box} models to understand their representations, this section focuses on highlighting the evidence supporting their decisions. We categorize these approaches into \textit{post-hoc} and \textit{inherently} interpretations.

\subsection{\textit{Post-hoc} Interpretation}
These methods, consider a pre-trained model and analyze how such a model process a textual input before producing a decision. When the model is black-box the interpretability method is \textit{model-agnostic}. If some assumptions are made on the architecture, the method is \textit{model-specific}.

\subsubsection{\textit{Model-agnostic} Explanations}\label{sec:NLPexplanations}
Given a black-box model $f: X \mapsto Y$ and an input $x$, the goal of \textit{model-agnostic} interpretation methods is to explain the individual prediction $f(x)$. The majority of these methods rely on perturbing $x$ according to some distribution $D_x$.

In \cite{ribeiro2016should}, Ribeiro et al. present LIME, one of the first state-of-the-art explainability algorithms, that approximates \textit{any} classifier or regressor locally with an interpretable model. LIME presents the interpretation $g$ for the user in terms of comprehensible explanations such as bag-of-words. Formally, LIME minimizes $\mathcal{L}(f,g,\pi) + \Omega(g)$, 
where $\mathcal{L}(f,g,\pi)$, a measure of how unfaithful the explanation $g$ is to the original model $f$ with the unfaithfulness is computed in a locality defined by a proximity measure $\pi$ and $\Omega(g)$, the complexity of the explanation $g$. The explanation is created by approximating $f$ \textit{locally} by an interpretable one. $\mathcal{L}(f,g,\pi)$ is approximated by sampling around an input $x$ according to $\pi_x$, performing perturbations on the input then explaining linearly, respecting thus local faithfulness. 

Later in 2018, Ribeiro et al. \cite{ribeiro2018anchors} argue that the coverage of the explanations generated by LIME is not clear. For instance {``not'', ``good''} could be an explanation of negative sentiment and {``not'', ``bad''} could be that of a positive one. Thus, the generated explanation does not clearly state when the word ``not'' has a positive/negative influence on the sentiment. Ribeiro et al. \cite{ribeiro2018anchors} define ``anchors'' as if-then rules to generate model agnostic-explanations. A perturbation is applied by replacing ``absent'' tokens with random words having the same POS tag drawn according to an embedding similarity-based probability distribution. Experiments on visual question answering, and textual and image classification show that the anchors model enhances the precision of explanations with less effort to understand and apply.

A landmark in the timeline of the ExAI method is achieved when Lundberg and Lee theoretically showed that many interpretability methods and metrics can be unified in one approach that exploits game theoretical concepts in \cite{lundberg2017unified}. Inspired by the Shapley value of game theory, SHAP values are proposed as a measure of the feature importance for a model's prediction. SHAP is proved to be a unified measure that different methods such as LIME \cite{ribeiro2016should}, Deep LIFT \cite{shrikumar2017learning} and layer-wise relevance propagation \cite{bach2015pixel} tried to approximate in the literature.
Later, Chen et al. \cite{chen2018shapley} extend SHAP and proposed the L-Shapley and C-Shapley measure by exploiting the underlying graph structure to reduce the number of model evaluations. In \cite{alvarez2017causal}, explanations of black-box models are formulated as groups of input-output tokens causally related. Explanations are generated by querying with perturbed inputs and solving a partitioning problem to select the relevant components. Variational auto-encoders are used to derive meaningful input perturbations.

We draw the reader's attention to the fact that the approaches discussed above \cite{ribeiro2016should,ribeiro2018anchors,lundberg2017unified,alvarez2017causal} are not exclusive to textual data. While other similar explainability methods exist \cite{strumbelj2010efficient,baehrens2010explain,lundberg2017unified}, they are not validated on NLP tasks, thus beyond the scope of this work.

\subsubsection{\textit{Model-specific} Explanations}
Instead of considering black-box models, K{\'a}d{\'a}r et al. \cite{kadar2017representation} interpret specific recurrent architectures in RNN models by quantifying the contribution of each input to the encoding of a GRU architecture. Their \textit{omission score} is computed by measuring the salience of each word $s_i$ in a sentence $s_{1:n}$ by observing how much the representation of the sentence when omitting the word $s_i$ would deviate. The results show the sensitivity to the information structure of a sentence and selective attention to lexical, semantic, and grammatical relations. Similarly, Arras et al. \cite{arras2019explaining} adapt the Layer-wise Relevance Propagation (LPR) method of \cite{bach2015pixel} to explain the predictions of accumulators and gated interactions in the LSTM architecture in particular.

\cite{kadar2017representation} and \cite{arras2019explaining} exploit general ExAI methods, such as perturbation and layer propagation, to explain NLP models. Although such methods do not require human annotation or intervention, they might be computationally expensive. Prior to that, researchers developed methods that imitate the thinking process of human beings by relying on human annotations to derive explanations. Those attempts are grouped under the  ``rationalization'' framework that aims at explaining why a specific instance belongs to a category by extracting a ``rationale'' along with the network annotation. According to \cite{lei2016rationalizing}, a ``rationale'' can be defined as ``sub-sets of the words from the input text that satisfy two key properties. First, the selected words represent short and coherent pieces of text (e.g., phrases) and, second, the selected words must alone suffice for prediction as a substitute of the original text.'' Rationales are proved to help to learn domain-invariant representations that can induce machine attention \cite{bao2018deriving}.

In \cite{zaidan2007using}, rationale-annotated data is exploited to aid learning by providing the learning algorithm (Support Vector Machines, SVMs, in this case) with hints as to which features of the input the algorithm should attend. Later, Zhang et al. \cite{zhang2016rationale}, extended the idea to CNNs by incorporating rationales for text classification in a Rationale-Augmented CNN (RA-CNN). RA-CNN computes a document-level vector representation by summing over its constituent sentence vectors weighted by the likelihood that the sentence is a rationale in support of the most likely class. 
Adversarial learning is utilized in \cite{yu2019rethinking} to improve the performance of rationale extraction so as not to leave any useful information out of the selection. The outcome is also incorporated in \cite{yu2019rethinking} into the selection process to improve the predictive accuracy through more comprehensive rationale extraction. Differentiable binary variables are introduced in \cite{bastings2019interpretable} to further improve the rationale extraction by augmenting the objective with re-parameterized gradient estimates. Game-theoretic concepts are incorporated in \cite{chang2020invariant} to derive a rationalization criterion that approximates finding causal features without highlighting spurious correlations between inputs and outputs.

A common benchmark to evaluate the rationalization attempts is presented in the ERASER framework of \cite{deyoung2020eraser}. Other approaches include, but are not limited to, contrastive textual explanations \cite{ross2020explaining}, representation erasure \cite{li2016understanding} and information theoretical measures \cite{chen2018learning}.

\subsection{\textit{Inherently} Interpretable Models}
While the previous interpretability methods work on pre-trained models, the methods discussed in this section develop models that can classify textual data while explaining their particular decisions. In other words, for a textual input $x$ the models discussed next can output a decision (class) $y$ and an explanation $E$ formulated differently in each reference. Using the terminology of this work, we further cluster these methods in \textit{model-agnostic} and \textit{model-specific} groups. 
\subsubsection{\textit{Model-agnostic} Explanations}
As an extension to the previously discussed ``rationalization'' attempts \cite{zaidan2007using,zhang2016rationale}, Lei et al.  \cite{lei2016rationalizing} propose an automated approach to extract rationales from the text as subsets of the text words that are coherent and short but lead to the same predictions. For this purpose, the authors train a generator and encoder that learn the classification as well as the explanation. While the former specifies a distribution over subsets of words as candidate rationales, and the latter processes them for prediction. This is achieved by (1) forcing the produced rationale, $z$ (set of words) to be sufficient as a replacement for the input text $x$ in predicting the output $y$, i.e. $||\textit{enc}(z,x) - y||$ small (2) while maintaining short, i.e. small $||z||$, and coherent rationales. Coherency is satisfied by encouraging words to form meaningful phrases (consecutive words) rather than sets of isolated words. Hence,  the sum of  $|z_t-z_{t-1}|$ is minimized. Then, doubly stochastic gradient descent is used to minimize the following objective over training instances of the form $(x,y)$: 
\begin{equation}
\sum_{(x,y)} \mathbb{E}_{z\approx \textit{gen}(x)} \bigg[
||\textit{enc}(z,x) - y||_2^2 + \lambda_1||z|| +\lambda_2\sum_t|z_t-z_{t-1}|
\bigg]
\end{equation}

\subsubsection{\textit{Model-specific} Explanations}

 Liu et al. \cite{liu2019towards} consider local explanations for particular inputs but present a generative explanation framework that learns to generate fine-grained explanations \textit{inherently}, while making classification decisions in a \textit{model-agnostic} manner. The prediction component is composed of a (1) text encoder that takes the input text sequence $S$ and encodes it in a representation vector $v_e$ and (2) a category predictor $P$ that outputs the category corresponding to $v_e$ along with its probability distribution. The explanation component consists of an explanation generator that takes $v_e$ and generates fine-grained explanations $e_c$, as a set of words that explain the model's decision. Fine-grained explanations can be thought of as attribute-specific rationales. In other words, in sentiment analysis applications, if a product has three attributes: quality, practicality, and price, the review ''the product has good quality with a low price'', can be explained by \textit{``low''} as a fine-grained explanation for the price and ``high'' as a fine-grained explanation for the quality. In \cite{liu2019towards}, those explanations $e_c$ are provided in two datasets collected by the authors of the work. 
 
 Training consists of minimizing a combination of two-loss factors: the classification loss and the explanation loss. To avoid cases where the generative explanations are independent of the predicted overall decision, the authors define an explanation factor that helps build stronger correlations between the explanations and the predictions. More specifically, a classifier $C$ is trained to predict the category from the explanations and not the original input text. $C$ is then used to provide more robust guidance for the text encoder to leverage the generation process by generating a more informative representation vector $v_e$. Experiments show that the explanation factor enhances the performance of the base predictor model. 

Reasoning over Knowledge Graph \cite{sap2019atomic} presents a promising way to explain NLP systems in a structured way in the form of $<source\_entity, relation, target\_entity>$ \cite{wang2020commonsense}. Graph reasoning has been addressed as a variational inference problem \cite{chen2018variational}, random-walk search \cite{gardner2014incorporating} and a Markov Decision Processes \cite{xiong2017deeppath,das2018go}. Recently, knowledge graph reasoning has attracted researchers in the NLP and ExAI communities. Readers are referred to \cite{chen2020review} where its basic concept, definitions, and methods are surveyed. 

\subsection{Discussion}
To the best of our knowledge, \textit{inherently} interpretable models for explaining individual decisions of NLP models are restricted to the work of \cite{liu2019towards,lei2016rationalizing}. This can be explained by the fact that current language models require large computational costs in terms of time and computational resources. Modifying the architecture to an explainable one and retraining the current high-performance models will be expensive.  


On the evaluation level, since \textit{model-agnostic} approaches were not specific to textual data, their evaluation was general and not specific to NLP cases. For instance, LIME's evaluation for textual data was done qualitatively by getting insights on a model trained to differentiate ``Christianity'' from ``Atheism'' on a
subset of the 20 newsgroup dataset  \cite{ribeiro2016should}. On the other hand, SHAP \cite{lundberg2017unified} is not explicitly tested on NLP tasks and \cite{ribeiro2018anchors}'s anchors were tested on tabular datasets. Although a common evaluation framework is not followed with the \textit{model-specific} approaches, textual data is used to evaluate the described explainability methods. This evaluation is inspired by sensitivity analysis where the sentence representation is monitored to evaluate how much deviation is observed when the words outputted by the explainability method are omitted \cite{kadar2017representation,arras2019explaining}. 
Regarding the \textit{inherently} interpretable models, one can clearly see that both methods \cite{liu2019towards,lei2016rationalizing} rely on the encoder-decoder design to generate explanations. \cite{lei2016rationalizing} evaluate their approach on
multi-aspect sentiment analysis and compare their explanations to manually annotated test cases. Although \cite{liu2019towards} have similar motivation, they do not compare to \cite{lei2016rationalizing}. The authors report instead a BLEU score for their generated explanations. They also report the performance of a classifier on the fine-grained explanation instead of the initial input and employ a human evaluation framework. 

%% file: future.tex
\begin{figure}[h]
    \centering

    \includegraphics[width=0.5\textwidth]{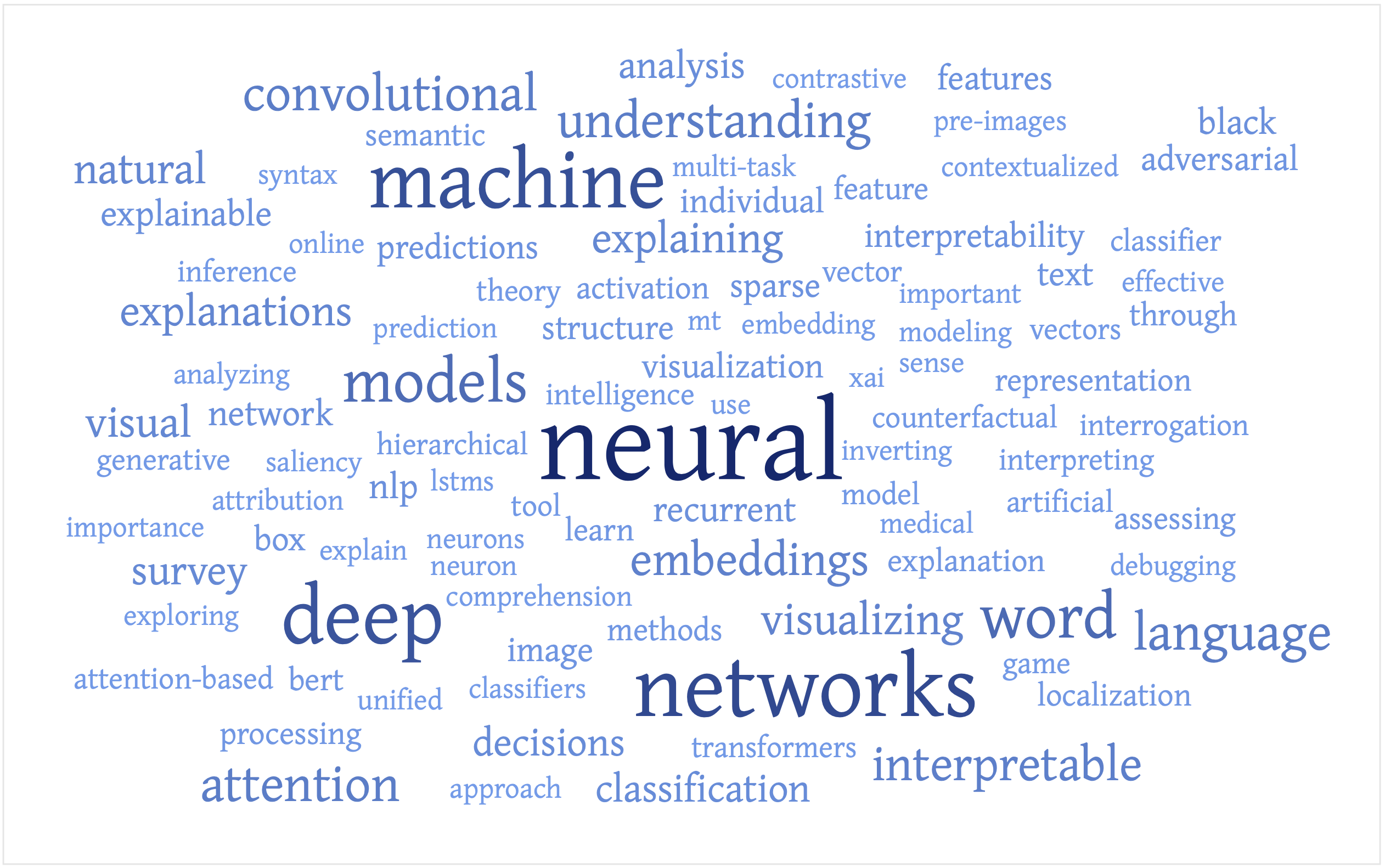}
    \caption{Most frequent terms used in the titles of the publications referenced in this work}
    \label{fig:word_cloud}
\end{figure}
\begin{figure}[h]
    \includegraphics[width=0.5\textwidth]{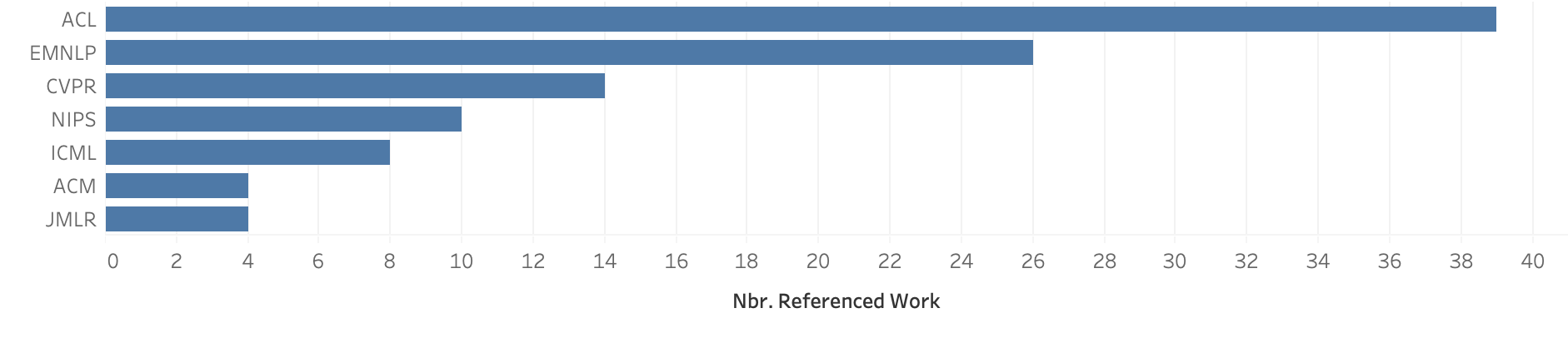}
    \caption{Top 7 conferences and journals referenced in this work}
    \label{fig:nbr_refs}
\end{figure}

To better understand the interest in ExAI in NLP, we consider the titles of the papers that we reference in this work and we visualize word frequencies in Figure~\ref{fig:word_cloud}. One can see that \textit{deep} \textit{networks}, \textit{representation} or \textit{embedding} methods and \textit{attention} models are very frequent in our referenced work. The importance of \textit{visualization}s or \textit{visual} clarification techniques in explaining deep models can be also inferred from the frequencies. The figure also hints at the fact that ExAI in NLP in the referenced work is mostly focused on \textit{understanding} the inner workings of the underlying models rather than understanding a particular output of \textit{classification}. Additionally, to reflect the interest in ExAI in general within the machine learning community, we show the top 7 conferences and journals referenced in this work in Figure~\ref{fig:nbr_refs}. The lack of journals that study or survey ExAI methods in general, and in NLP, in particular, is reflected in the figure. Researchers are publishing their ExAI methods and discoveries in general AI conferences or conferences that focus on linguistics and natural language.


\begin{figure}[h]
    \centering
    \includegraphics[width=0.5\textwidth]{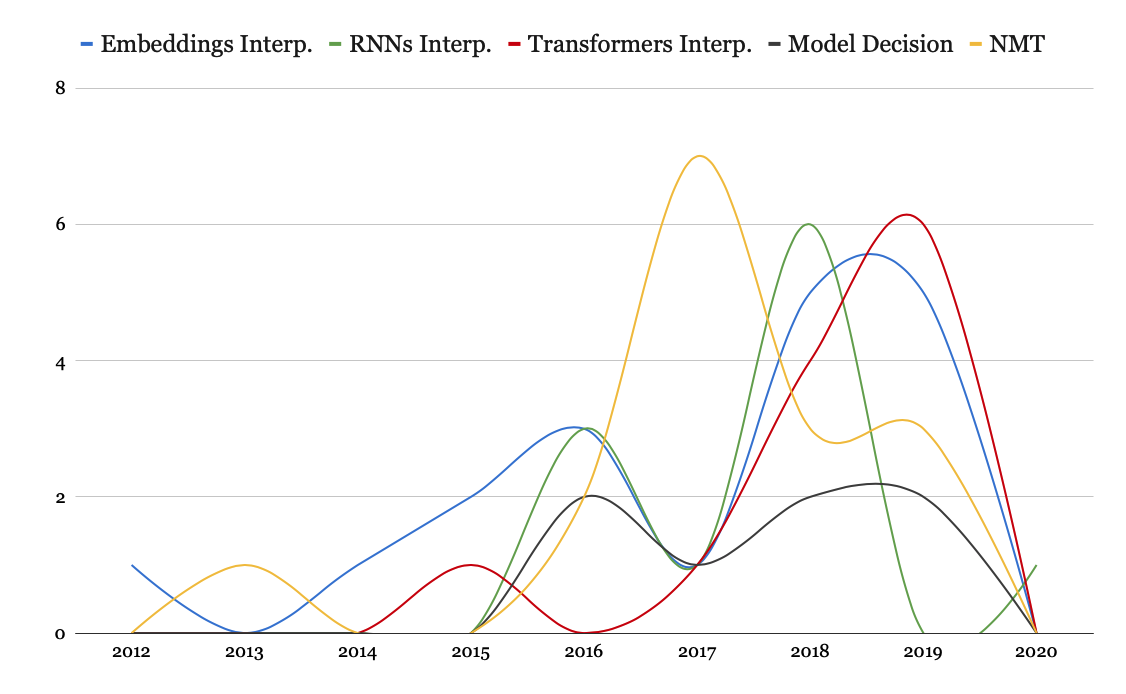}
    \caption{Timeline for ExAI in NLP}
    \label{fig:timeline}
\end{figure}

Furthermore, Figure~\ref{fig:timeline} shows how the research interest in some ExAI categories is progressing since 2012. For instance, The interpretability of word embeddings has attracted researchers since 2015, when the concern about bias in the machine learning model has emerged after Google Photos application tagged a black woman as Gorilla \cite{google_black}. The introduction of transformers in 2017 has also encouraged researchers to study the magic behind their state-of-the-art performance on different NLP tasks. Moreover, the unprecedented breakthroughs in NMT in 2016 \cite{wu2016google} encouraged research work on ExAI applied to NMT models afterward.

Although recent years are witnessing significant growth in ExAI methods applied to NLP models, these methods are not fulfilling their potential yet. For instance, the evaluation of ExAI methods is lacking a unique testing framework where metrics and datasets are well-designed. In the assessment of ExAI's current methods, different NLP tasks such as POS tagging, word intrusion, and correlation experiments are explored. However, these tasks are paper-specific and they are not used to compare different methods. This brings to the front the need for a common evaluation framework where human-labeled datasets are well defined and NLP tasks are described within the syntactic or semantic aspect that they aim to explain. Additionally, an area that is yet to be explored by researchers is the quantitative assessment of the interpretability of the embedding spaces and the effectiveness of each dimension in semantically and syntactically encoding a particular concept.

Moreover, the majority of the work studied in this survey post-process NLP models to explain their decisions and inner workings. \textit{Inherently} interpretable NLP models are under-examined due to the fact that retraining a model after modifying its architecture is very expensive and might not achieve the same performance. Exploring explainable designs for NLP models can thus be one of the main subjects for future work. Another direction for future work is the level at which explanations for language models are provided. So far, the explanations are provided as either the contribution of individual words to the decision or the layer/neuron at which syntax or semantics are encoded. However, text analysis is a multi-step process: after extracting information from textual unstructured data, analysis is applied to the extracted information to reach knowledge. Wisdom and logic come at the highest level of semantics. Thus, explanations provided on the individual input words will be ignoring the hierarchy of the text understanding process which will affect their efficacy. 

%% file: conc.tex
This work presents the first comprehensive survey on explainability methods in the NLP field that combines ExAI methods on the input-, processing- and output levels. According to the assumptions made on explained models, we make the distinction between \textit{model-agnostic} and \textit{model-specific} methods. We further distinguish between explanation in a \textit{post-hoc} manner and explanation that results in \textit{inherently} interpretable models.

We present different attempts to interpret word embedding models that are recently serving as inputs to almost every NLP network. Some of those methods rely on altering the embedding space by imposing a sparsity constraint or applying a rotation transformation while others integrate external knowledge bases and ontologies or rely on bidirectional language models to derive contextualized embeddings. Moreover, we survey existing work on the interpretation of hidden representations of NLP models, general RNNs, and transformers in terms of human-understandable concepts. We discuss the debate over the interpretability of attention weights and we derive the following conclusion: attention weights are relatively more \textit{inherently} interpretable than traditional models' parameters but further analysis is required for complete transparency in attention-based models. Additionally, we present the research work that explains a particular model decision inspired by general ExAI methods or designed specifically for NLP models. We also discuss different visualization platforms that present user-friendly explainability schemes based on perturbations and attention weights. 

\begin{figure}[h]
    \centering
    \includegraphics[width=0.9\textwidth]{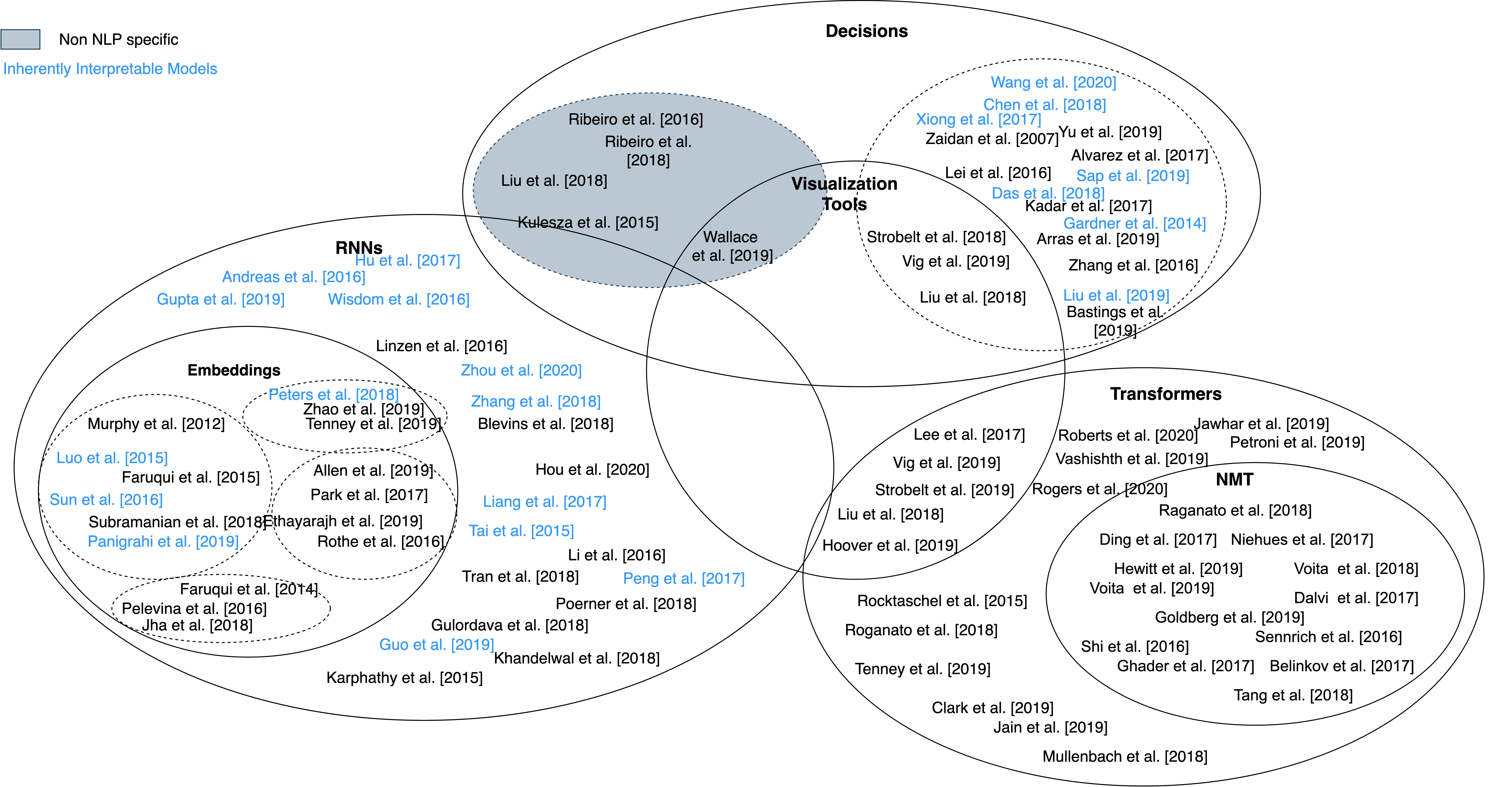}
    \caption{References for Sections~\ref{sec:NLPembeddings}, \ref{sec:RNNs}, \ref{sec:transformers} visualized over similarity of scopes. }
    \label{fig:venn_full}
\end{figure}

Figure~\ref{fig:venn_full} summarizes the work done on the interpretability of word embeddings, inner workings of RNNs and transformers, the model's decision, and the different visualization methods while highlighting the interconnections between the different methods.

To date, there is no common evaluation ground for explainability methods on NLP models. In this work, we shed the light on different empirical setups, datasets, and metrics that can be designed to assess the performance of ExAI in NLP. These setups are aggregated according to \textit{what} the corresponding ExAI method is addressing: embeddings, inner workings, or model's decisions. We further examine these evaluation methods to discriminate between evaluations that explicitly assess the interpretability and those that are of qualitative nature or need further analysis to extract insights useful from an explainability perspective.  Finally, we discuss the limitations of existing ExAI methods while highlighting research areas that can possibly be the focus of researchers in their future exploration.


%% file: nmt.tex
Translation is a particular NLP application that attracted interest from researchers to study the internal dynamics of its deep models. In this section, we present the different attempts at interpreting the general deep models and specific transformers that are developed to serve NMT applications.

Deep learning approaches for NMT are mostly encoder-decoder deep networks that are trained, end-to-end, on sentence pairs \cite{cho2014learning,sutskever2014sequence}. Instead of integrating syntax-based rules on the source and/or target languages, such sequence-to-sequence models elegantly learn to encode the source sentence in a high-dimensional vector and then decode it into the target sentence. It is notoriously hard to interpret the procedure that encodes syntactic and semantic information about the source/target languages. In their article on the challenges that face NMT tasks, Koehn and Knowles \cite{koehn2017six} shed the light on the importance of developing a better understanding of the internal representations of such models in order to achieve more robust models. 

Recently, researchers have been investigating \textit{how} and \textit{where} syntactic and semantic information has been learned. Such approaches aim at investigating the NMT deep learning models to gain better insights into how well these models encode language which could facilitate using semantic information for improving NMT systems.

Recent research work suggests that NMT transformers are able to encode syntactic information such as dependency parse trees \cite{hewitt2019structural,raganato2018analysis} and subject-verb pairings \cite{goldberg2019assessing}. Prior to that, Ding et al. \cite{ding2017visualizing} use the layer-wise propagation technique \cite{bach2015pixel}, previously used in general-purpose explainability methods, to visualize and better understand the NMT model. The approach back-propagates relevance along with the attention network recursively to the contribution of each contextual word to the hidden states of the attention-based encoder-decoder model. The resulting visualization is shown to offer interesting insights into relevant words in the NMT task as well as the causes of the translation errors.

Moreover, Tang et al. \cite{tang2018analysis} and Ghader et al. \cite{ghader2017does} focused on analyzing attention mechanisms in NMT. The former work compares attention to alignment and studies whether attention is able to model translational equivalent or to capture more information. By measuring (1) the cross-entropy between attention and soft alignment as a loss function and (2) the concentration of attention as the entropy of the attention distribution, the authors showed that attention agrees with alignment to a high degree. However, this agreement is substantially dependent on the attention mechanism and the POS tag of the word being generated. Moreover, attention is shown to capture higher-level information rather than just the translational equivalent of a word. Ghader et al. \cite{ghader2017does} focus on the word sense disambiguation in NMT, where translations are evaluated on the ambiguous nouns directly and \textit{fast-align} \cite{dyer2013simple} is used to align the translation of the ambiguous words. The results show that attention is more likely to be distributed to the ambiguous noun itself rather than its context, which suggests that the hidden states of the NMT encode the contextual information necessary for disambiguation. More interestingly, in transformers, the first few layers are responsible for gradually learning the alignment while the last layers extract relevant features from the context tokens.

Shi et al. \cite{shi2016does} focus on the semantic information in NMT models and try to understand \textit{where} and \textit{how much} of syntactic information does the encoder learn about the source language. For this purpose, two methods have been proposed. The first one considers the representation vectors learned by the encoder of sentences (for sentence-level labels) and individual words (for word-level labels) and tries to predict the syntactic labels of individual words in the source sentence. The high accuracy of the syntactic labels predictor shows that the encoder is indeed able to capture significant sentence-level syntactic information as well as word-level information where words that have similar POS and Smallest Phrase Constituents (SPC) labels tend to be clustered correctly. When considering different cell states, experiments show that local features (such as POS and SPC) are encoded in the lower layers, while deeper layers tend to encode more abstract and global syntactic information.  The second method relies on a retrained linearized-tree decoder to extract the constituency tree of the source sentence from the representation vector learned by the encoder. The parse tree is then analyzed to show that the considered sentence vectors are able to encode the syntax but can still miss some subtle syntactic details.  

Similarly, Belinkov et al. \cite{belinkov2017evaluating} consider different layers in an NMT model and evaluated the hidden representations on semantic and POS tagging tasks. In addition to showing that higher layers encode a more global perspective, the authors show that when the encoder-decoder is trained on English-English, poorer representations for POS and semantic tagging are produced. Their work highlights the importance of the translation component in learning useful syntactic representations. Moreover, the target language is shown to have little effect on the encoded information. A parallel work for Belinkov et al. \cite{belinkov2017neural} assesses the ability of the representations in NMT systems at different granularity levels of granularity in learning morphological structures. Through extrinsic POS and morphological tagging tasks, Belinkov et al. were able to conclude that lower layers are able to encode word morphology better while deeper ones capture semantics. After all, morphology is better learned by character-based representations rather than word-based ones. 

\cite{niehues2017exploiting} leverage multi-task learning to improve NMT performance by jointly training several NLP tasks in one system to introduce external knowledge into an end-to-end NMT model. \cite{voita2018context} compare the hidden representation of their context-aware NMT system to human-annotated co-reference relations to show that their model implicitly encodes anaphora. \cite{sennrich2016grammatical} introduces translation errors to the contrastive translations to explain the trade-off between the translation generalization to unknown words where character-level decoders perform better and modeling morpho-syntactic agreement, where long-term information needs to be carried and where sub-word models yield enhanced performances. \cite{dalvi2017understanding} define a quantitative measure of how well the NMT model, the decoder specifically, is able to learn morphological features as the accuracy of a classifier that is trained to predict the morphological tag of features extracted from the NMT model. The experiments show that the decoder learns limited morphological structures, which can be explained by the fact that the encoder and the attention mechanism help the decoder generate the needed morphological forms. 

Recently, Voita et al. \cite{voita2019analyzing} focused on the individual contribution of every attention head to the overall performance of the task. For this purpose, layer-wise relevance propagation was used to quantify the degree to which different heads at each layer contribute to the model's prediction, where heads with higher relevance value may be more important. The relevance values were then compared to the ``confidence'' of each head, which is defined as the average of its maximum attention weight, excluding the end of sentence symbol. Consequently, a confident head is the one that attends, with a high proportion, to a single token and is intuitively considered to be important to the translation task.  Experiments show that relevance values agree with the computed ``confidence''. Additionally, to interpret the role of attention heads, the authors identified three main roles: (1) positional where the head attends to adjacent tokens, (2) syntactic where it points to related tokens according to a specific syntactic relation and (3) rare words where the head points to the least occurrent token. The criteria based on which the authors differentiate between the three types assume that: (1) A head is positional if its maximum weight is assigned to an adjacent token at least 90\% of the time. (2) For each head, the ``accuracy'' is computed as the portion of time it assigns its maximum attention weight to a token with which there exists a dependency relation according to \cite{manning2014stanford}. The ``accuracy'' of a baseline model is computed as the most frequent relative position for the particular dependency relation. A head is syntactic if its ``accuracy'' is at least 10\% higher than that of the baseline model. (3) A head is rare if it points to the least frequent words in the sentence. The results reveal that ``accuracy'' helps detect heads that learn syntactic relations with accuracies that are higher than the baseline. Consequently, the encoder is indeed able to perform syntactic disambiguation of the input. Moreover, more than one head could be responsible for the same dependency relation. More importantly, the first layer contains a head that points to rare words and this head is judged to be the most important to the model's predictions.

Table~\ref{tbl:nmt} summarizes the aspects of interpretability studied within the NMT framework while highlighting the method used and the findings reached by each work in terms of interpreting the knowledge encoded in NMT models. 

\begin{table}[h]
\small
\begin{tabular}{p{1cm}p{0.8cm}p{3.5cm}p{3.5cm}p{5cm}}
\hline
 &
  \textbf{Ref} &
  \textbf{Aspect} &
  \textbf{Method} &
  \textbf{Findings} \\ \hline\hline
\multirow{2}{*}{\textbf{2016}} &
  \cite{shi2016does} &
  Encoding of syntactic information &
  Syntactic label prediction and linearized -tree encoder &
  The ability of the encoder to learn syntactic information at different levels \\ 
 &
  \cite{sennrich2016grammatical} &
  The trade-off between generalization and morpho-syntactic agreement modeling&
  Introduction of translation errors in contrastive explanations &
  (1) Better generalization of character-level decoders and (2) better performance of sub-word models on morpho-syntactic agreement tasks \\ \hline
\multirow{5}{*}{\textbf{2017}} &

  \cite{ghader2017does} &
  Word sense disambiguiation &
  Fast align \cite{dyer2013simple} &
  Encoding of context needed for sense disambiguation in hidden states \\ 
 &
  \cite{belinkov2017evaluating},\cite{belinkov2017neural} &
  POS encoding &
  Evaluation of hidden representation on POS tagging tasks &
  Better encoding of morphology in higher layers and (2) negligible effect of the target languages on the encoding. \\  
 &
  \cite{niehues2017exploiting} &
  Integrate of external knowledge &
  Multi-task training &
  Enhancement of the performance of NMT models by the multi-task learning\\ 
 &
  \cite{dalvi2017understanding} &
  Morphology of learned features &
  Accuracy on morphological tags prediction &
  Limited morphological structures learned by the decoder \\  
 &
  \cite{ding2017visualizing} &
  Contribution of words in the hidden layers &
  Layer-wise relevance propagation \cite{bach2015pixel} &
  Causes of translation errors and relevance of words \\ \hline
\textbf{2017-2018} &

  \cite{ghader2017does}, \cite{tang2018analysis} &
  Ability of attention to model equivalence &
  Cross-entropy and concentration of entropy &
  Agreement between attention and alignment (equivalence) \\ \hline
\textbf{2018} &

  \cite{voita2018context} &
  Anaphora encoding &
  Comparison to human-annotated coreference &
  The ability of NMT models to encode anaphora \\ \hline
\textbf{2019} &

  \cite{voita2019analyzing} &
  Contribution of attention heads &
  Layer-wise relevance propagation \cite{bach2015pixel} &
  (1) Syntactic disambiguation by encoders and more than one head can encode dependencies \\ \hline
\end{tabular}
\caption{Summary of the findings of the ExAI methods in NMT along with the interpretability aspect and the used methods.}\label{tbl:nmt}
\end{table}

%% file: vis.tex
The discussed explainability methods are mostly static analyses of NLP models that can lead to targeted insights into the language model but do not allow for dynamic interactions with the model. Interactive tools instead can dynamically help humans and experts better understand the model's internal reasoning by manually perturbing input and observing their effect on the behavior of the model.

\cite{kulesza2015principles} presented \textit{Explanatory Debugging} where the human-in-the-loop system outputs its candidate explanations to the users, who suggest necessary corrections back to the learning system. NLIZE \cite{liu2018nlize} is another tool that aids in the interpretation of NLP models through a perturbation-driven approach. Instead of only perturbing specific locations in the input, NLIZE also perturbs the hidden and the prediction-level representation to explain particular decisions.

In \cite{wallace2019allennlp},
Wallace et al. introduced \textit{AllenNLP Interpret}, 
a flexible framework for interpreting \textit{any} NLP models. \textit{AllenNLP Interpret} provides interpretation for primitives NLP models such as masked language modeling using BERT and reading comprehension, 
a set of interpretation methods such as saliency maps and adversarial attacks, along with a library of visualization components. Other techniques visualize attention weights as in \cite{lee2017interactive,liu2018visual,strobelt2018s,vig2019visualizing,hoover2019exbert} 
for transformers and sequence-to-sequence models.  While the interpretability of attention weights is debatable, these tools enable the understanding of the parts that deep networks are attending to in a visually appealing manner.

\begin{figure}[h]
    \centering
    \begin{subfigure}[t]{0.9\textwidth}
    \includegraphics[width=\textwidth]{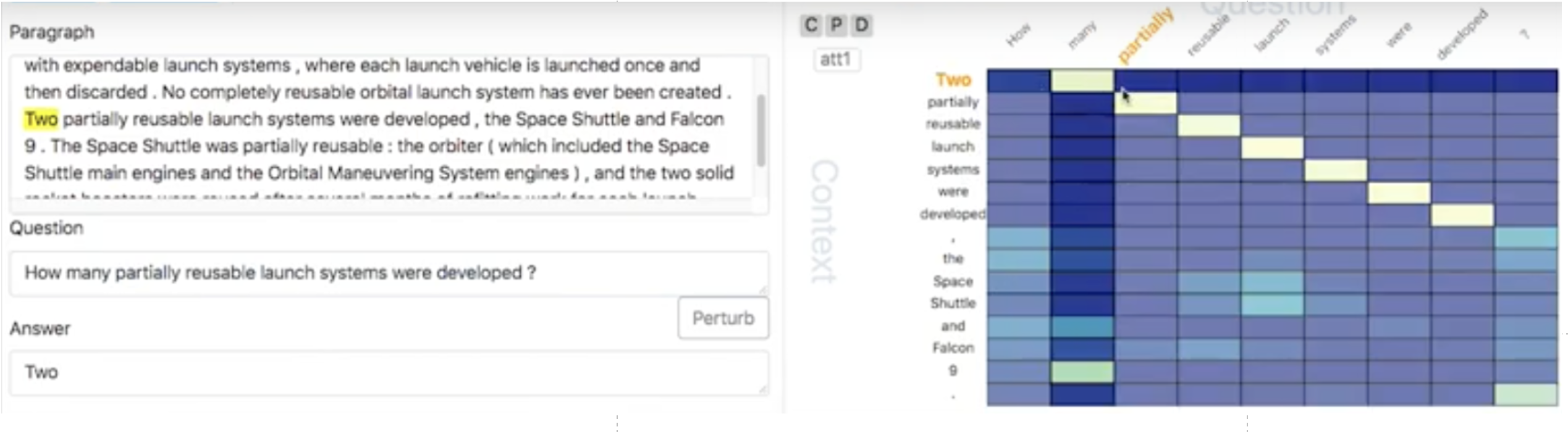}
    \caption{NLIZE \cite{liu2018nlize} explanations for a question answering task}
    \end{subfigure}
    
    \begin{subfigure}[t]{0.9\textwidth}
    \includegraphics[width=\textwidth]{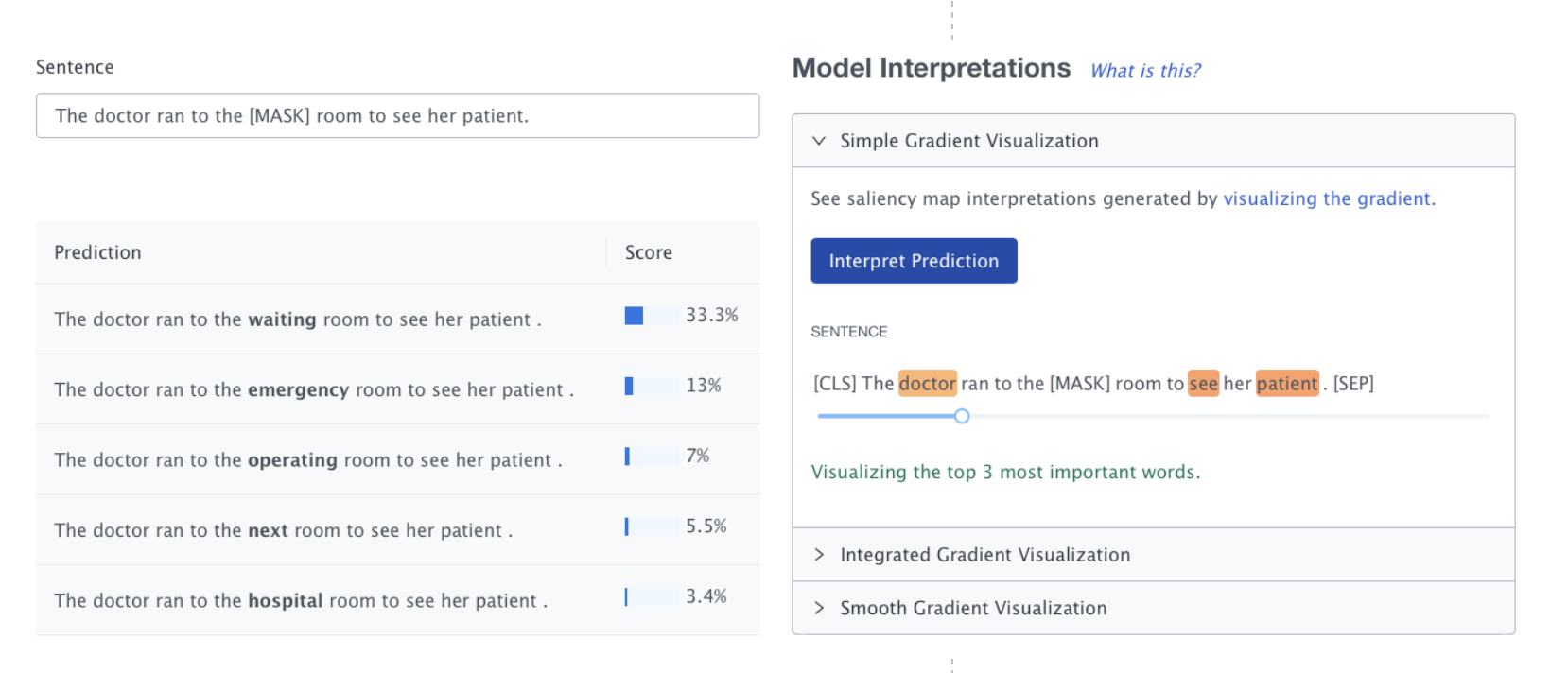}
    \caption{AllenNLP \cite{wallace2019allennlp} explanations for the word prediction tasks }
    \end{subfigure}
    
    \begin{subfigure}[t]{0.9\textwidth}
    \includegraphics[width=\textwidth]{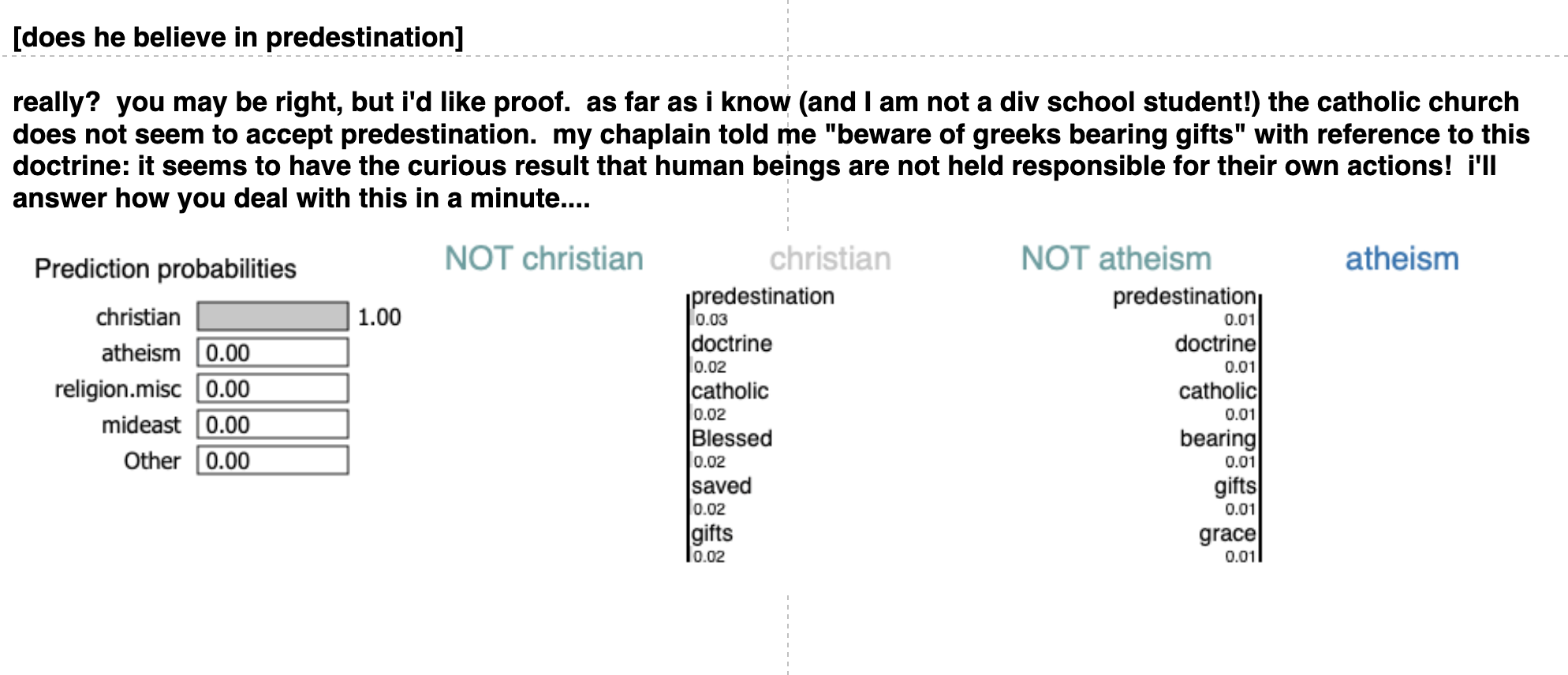}
    \caption{LIME \cite{ribeiro2016should} explanations on the news group classification task}
    \end{subfigure}    
    \caption{Screenshots of the user interaction with some of the visual explanation tools referenced in this work}
    \label{fig:vis_tools}
\end{figure}

IBM research also introduced their AI Explainability 360 tool in \cite{aix360}. Additionally, Ribeiro et al. \cite{ribeiro2016should} and Lundberg et al. \cite{lundberg2017unified} have also developed comprehensive visualization tools for their LIME and SHAP methods under python. The tools discussed in this section have been tested for the sake of this work, and screenshots of the user interaction are shown in Figure~\ref{fig:vis_tools}.